\documentclass{article}

\usepackage{microtype}
\usepackage{graphicx}
\usepackage{subcaption}
\usepackage{booktabs} 

\usepackage{hyperref}

\usepackage[preprint]{icml2026}

\usepackage{amsmath}
\usepackage{amssymb}
\usepackage{mathtools}
\usepackage{amsthm}
\usepackage{xspace}
\usepackage{enumitem}

\usepackage{multirow}
\usepackage{colortbl}

\usepackage[capitalize,noabbrev]{cleveref}

\newcommand{\toolname}{\mbox{\textit{UnCLE}}\xspace}

\def\cov{\mathrm{cov}}
\def\prec{\mathrm{prec}}
\def\AOPC{\mathrm{AOPC}}

\usepackage{amsmath,amsfonts,bm}

\def\eqref#1{equation~\ref{#1}}

\def\1{\bm{1}}

\def\vb{{\bm{b}}}

\def\vx{{\bm{x}}}

\def\vz{{\bm{z}}}

\DeclareMathAlphabet{\mathsfit}{\encodingdefault}{\sfdefault}{m}{sl}
\SetMathAlphabet{\mathsfit}{bold}{\encodingdefault}{\sfdefault}{bx}{n}

\def\sA{{\mathbb{A}}}
\def\sB{{\mathbb{B}}}
\def\sC{{\mathbb{C}}}
\def\sD{{\mathbb{D}}}

\def\sP{{\mathbb{P}}}
\def\sQ{{\mathbb{Q}}}
\def\sR{{\mathbb{R}}}

\def\sT{{\mathbb{T}}}

\def\sX{{\mathbb{X}}}

\newcommand{\E}{\mathbb{E}}

\newcommand{\R}{\mathbb{R}}

\theoremstyle{plain}

\theoremstyle{definition}

\theoremstyle{remark}

\usepackage[textsize=tiny]{todonotes}

\icmltitlerunning{Beyond Attribution: Unified Concept-Level Explanations}

\begin{document}

\twocolumn[
  \icmltitle{Beyond Attribution: Unified Concept-Level Explanations}



  \icmlsetsymbol{equal}{*}

  \begin{icmlauthorlist}
    \icmlauthor{Junhao Liu}{pku,klab}
    \icmlauthor{Haonan Yu}{pku,klab}
    \icmlauthor{Xin Zhang}{pku,klab}
  \end{icmlauthorlist}

  \icmlaffiliation{pku}{School of Computer Science, Peking University, Beijing 100871, China}
  \icmlaffiliation{klab}{Key Lab of High Confidence Software Technologies (Peking University), Ministry of Education}

  \icmlcorrespondingauthor{Xin Zhang}{xin@pku.edu.cn}

  \icmlkeywords{Machine Learning, ICML}

  \vskip 0.3in
]



\printAffiliationsAndNotice{}  

\begin{abstract}

There is an increasing need to integrate model-agnostic explanation techniques with concept-based approaches, as the former can explain models across different architectures while the latter makes explanations more faithful and understandable to end-users.
However, existing concept-based model-agnostic explanation methods are limited in scope, mainly focusing on attribution-based explanations while neglecting diverse forms like sufficient conditions and counterfactuals, thus narrowing their utility.
To bridge this gap, we propose a general framework \toolname to elevate existing local model-agnostic techniques to provide concept-based explanations.
Our key insight is that we can uniformly extend existing local model-agnostic methods to provide unified concept-based explanations with large pre-trained model perturbation.
We have instantiated \toolname to provide concept-based explanations in three forms: attributions, sufficient conditions, and counterfactuals, and applied it to popular text, image, and multimodal models.
Our evaluation results demonstrate that \toolname provides explanations more faithful than state-of-the-art concept-based explanation methods, and provides richer explanation forms that satisfy various user needs.

\end{abstract}

\section{Introduction}
\label{sec:intro}


The widespread application of machine learning models has created a strong demand for explanation methods.
In particular,
as the structure of machine learning models varies and evolves rapidly, and effective closed-source models (e.g., GPT-4~\citep{gpt4} and Gemini~\citep{geminiteam2024geminifamilyhighlycapable}) become more prevalent,
model-agnostic explanation methods show their appeal~\citep{survey_model_agnostic}.
These methods treat target models as black boxes and thus do not require access to internal model information, and provide explanations in forms including attributions, sufficient conditions, and counterfactuals, satisfying various user needs~\citep{survey_model_agnostic}.
On the other hand, providing concept-based explanations has emerged as a promising direction~\citep{concept_survey, EAC}, which are well-known to be more faithful and understandable to end-users~\citep{Zhang_Madumal_Miller_Ehinger_Rubinstein_2021, Ghorbani_Wexler_Zou_Kim_2019,Kim_Wattenberg_Gilmer_Cai_Wexler_Viegas_2018,EAC}.
As a result, there is a growing demand for providing concept-based model-agnostic explanations for machine learning models.

%

While providing concept-based model-agnostic explanations is promising, existing methods remain limited in scope. They are largely restricted to attribution forms~\citep{concept_survey} and lack support for richer explanation types such as sufficient conditions and counterfactuals. On the other hand, feature-level model-agnostic methods support diverse forms other than attributions, but they lack the interpretability and fidelity brought by using high-level concepts.

\begin{figure*}[ht]
    \centering
    \resizebox{0.85\textwidth}{!}{
        \includegraphics[]{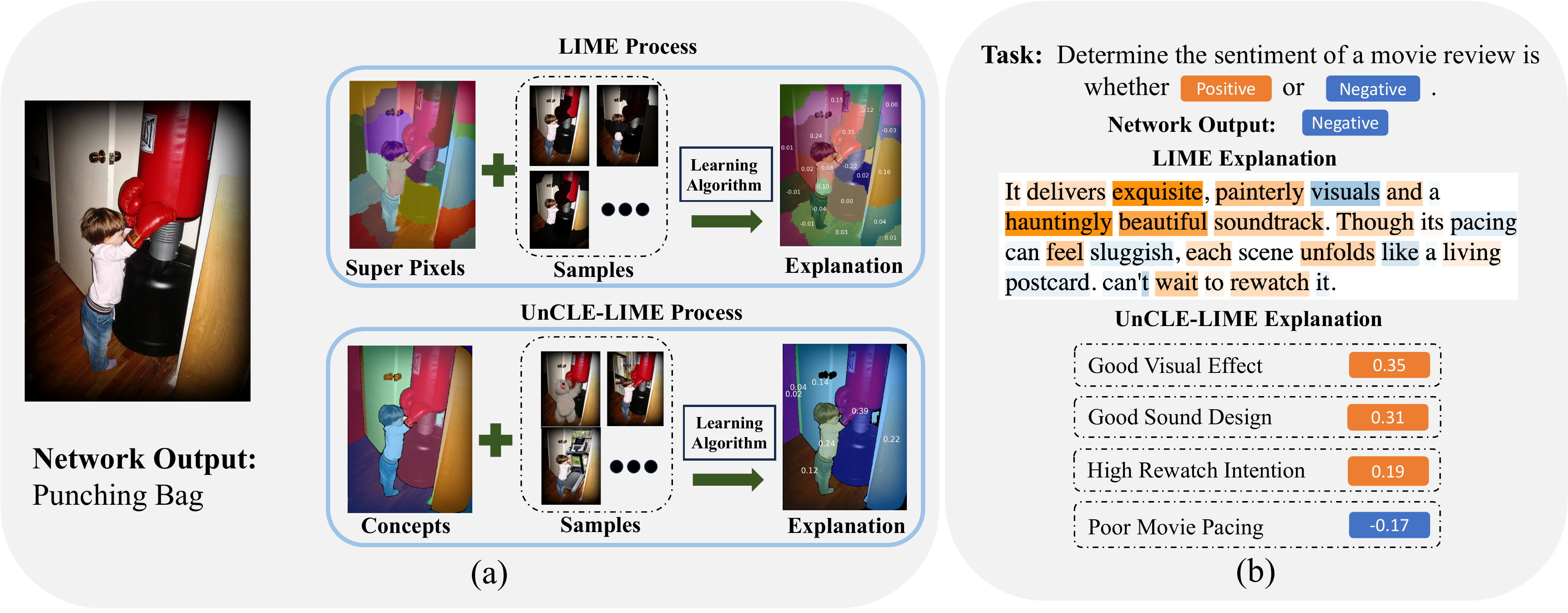} 
    }
    \caption{
Examples of using LIME and \toolname-augmented LIME to explain (a) an image classification model (YOLOv8) and (b) a text classification model (BERT).
The \toolname-augmented versions provide concept-based explanations, utilizing detected objects or topics rather than fragmented superpixels or words.}

    \label{fig:example-image-perturbation}
\end{figure*}

\begin{figure}[t]
    \centering
    \resizebox{\linewidth}{!}{
  \includegraphics[scale=0.33]{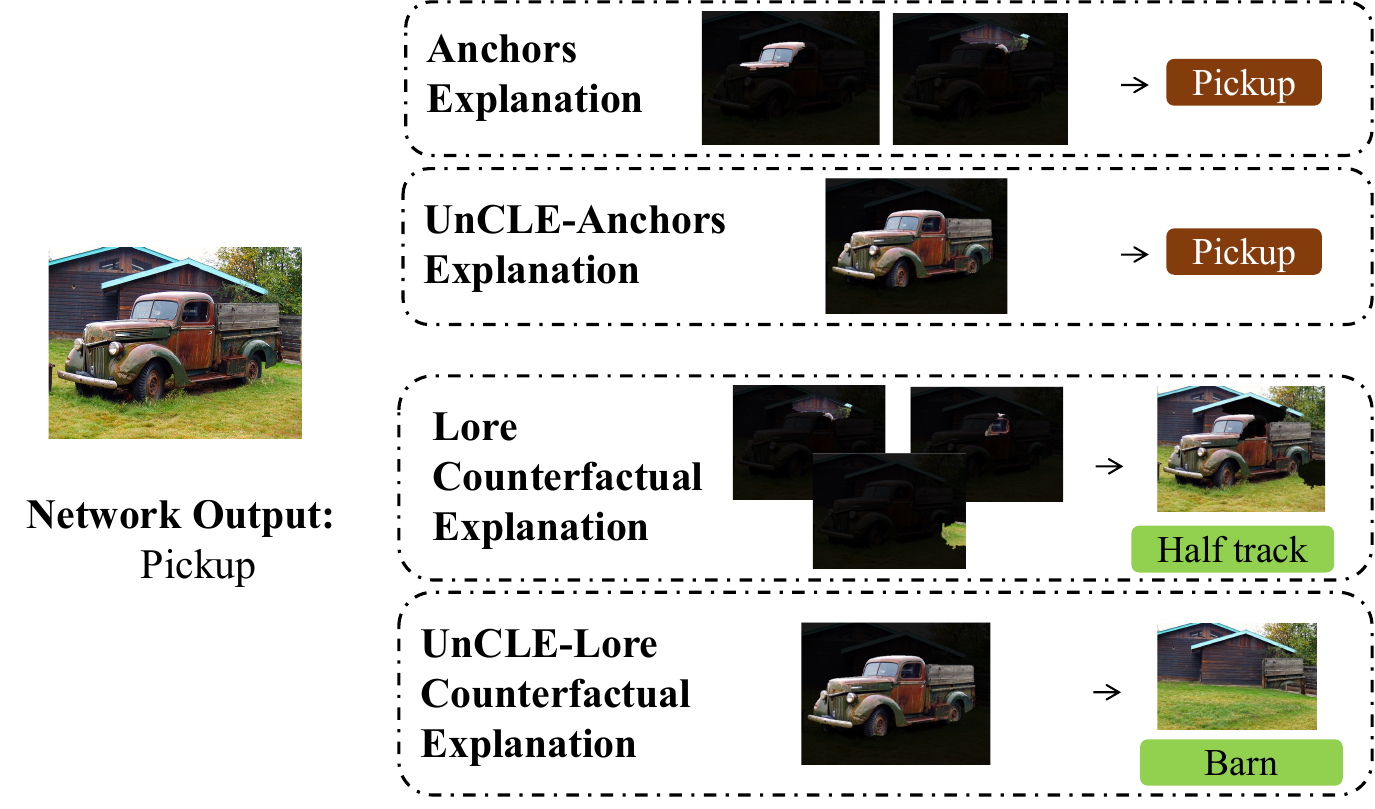} 
    }
    \caption{
    Example explanations from Anchors, LORE, and their \toolname-augmented versions.
    The Anchors explanation states that the presence of specific image regions guarantees that the model classifies the image as a pickup.
    The LORE explanation shows that masking these regions would lead the model to predict a different class.
    The \toolname-augmented versions provide concept-based explanations, using detected objects rather than fragmented superpixels.
    }
    \label{fig:example-image1}
\end{figure}
To bridge this gap, we propose \toolname, a general and lightweight framework to elevate existing feature-level model-agnostic methods from feature level to concept level, providing concept-based explanations in various forms beyond attributions, as well as concept-based attribution explanations of higher fidelity.
Additionally, we focus on local explanations, as they are tractable for end-users when explaining complex models used in real-world applications.
%

We achieve this goal by making two key observations.
First, elevating existing local model-agnostic methods to concept level does not require changing their core algorithms.
Since these methods follow a common workflow~\citep{ReX}, we only need to 1) extract high-level concepts from input data and 2) perform perturbations at concept level.

Second, given that several effective concept extraction methods already exist~\citep{ACE, ConceptGlassbox, TBM}, the only challenge is to perform perturbations at the concept level.
To this end, we propose using large pre-trained models to handle concept-level perturbations.

%

Figure~\ref{fig:example-image-perturbation} shows examples of using \toolname to elevate LIME~\citep{LIME} to provide concept-based explanations for image and text classification models.
%
Let us delve into the details of the image classification example.
%
Unlike vanilla LIME using fragmented superpixels, \toolname-augmented LIME generates explanations based on high-level concepts following \citet{EAC}.
%
In the perturbation phase, instead of masking fragmented superpixels, \toolname-augmented LIME directly modifies the detected objects in the image.
With this augmentation, the resulting attributions are grounded in semantically meaningful objects rather than low-level superpixels, making the explanations more interpretable for end-users.


%
%
%
%
%

Besides attribution-based methods, \toolname can also extend local model-agnostic methods of various forms to provide concept-based explanations, satisfying diverse user needs and offering a more comprehensive understanding of target models.
%
%
%
Figure~\ref{fig:example-image1} shows that \toolname can extend Anchors~\citep{Anchors} and LORE~\citep{lore} to provide concept-level rule-based sufficient conditions and counterfactual explanations, respectively. 
As \toolname provides all these explanations in a unified manner, users can obtain their desired explanations by simply selecting the explanation type they need.

Moreover, although \toolname is simple in design, it outperforms more elaborate, specially designed concept-based approaches, as our empirical evaluation results will later show.

In our evaluation, we first verified that large pre-trained models can faithfully perform concept-level perturbations.
Then we evaluated \toolname on explaining two text classification models, three image classification models, and one multimodal model.
%
Our evaluation results show that \toolname improves the fidelity of Anchors, LIME, LORE, and Kernel SHAP~\citep{SHAP} explanations by 56.8\%, on average, and \toolname-augmented methods outperform state-of-the-art concept-based explanation methods specifically designed for text models~\citep{TBM,yu2024latent} and image models~\citep{EAC}, respectively.
We also ran a human evaluation, which shows that \toolname helps users better use the explanations in downstream tasks.

\textbf{Contributions\;} Our contributions are as follows:
\begin{itemize}[topsep=0pt,parsep=0pt,partopsep=0pt, leftmargin=10pt]
    \item We introduce \toolname, a general and lightweight framework that elevates existing local model-agnostic explanation methods to concept level with minimal user effort, providing concept-based explanations with state-of-the-art performance, and in various forms beyond attributions.
    \item We propose using large pre-trained models to perform concept-level perturbations, enabling model-agnostic methods to generate concept-based explanations, and empirically verify the fidelity of these perturbations.
    \item We instantiate \toolname on four popular explanation methods: LIME, Kernel SHAP, Anchors, and LORE, and demonstrate its effectiveness through extensive experiments and human evaluations. \toolname achieves state-of-the-art performance in generating concept-based explanations. Additionally, except the concepts we used in Section~\ref{sec:intro} and ~\ref{sec:eval}, \toolname is flexible to work with various concepts, and we have shown examples in Appendix~\ref{app:more-concept}.
\end{itemize}

\section{Background and Related Work}
\label{sec:pre}
Our work is related to model-agnostic explanation methods and concept-based explanation methods.
We next introduce the background knowledge and related work.

\subsection{Machine Learning Models}
We consider a machine learning model as a function \(f\) that maps an input vector \(\vx\) to an output scalar \(f(\vx)\).
Formally, let \(f: \sX \rightarrow \R\), where \(\sX=\sR^n\).
%
%
Let \(\vx_i\) denote the \(i\)-th feature value of \(\vx\).

\subsection{Local Model-Agnostic Explanation Methods}
A local model-agnostic explanation method \(t\) takes a model \(f\) and an input \(\vx\), and generates a local explanation \(g_{f,\vx}\) to describe the behavior of \(f\) around \(\vx\), i.e., \(g_{f,\vx} := t(f, \vx)\).
An explanation \(g_{f,\vx}\) (\(g\) for short) is an expression formed with predicates. Each predicate \(p\) maps an input \(\vx\) to a binary value, i.e., \(p: \sX \rightarrow \{0, 1\}\), indicating whether \(\vx\) satisfies a specific condition.

Existing local model-agnostic explanation methods follow a similar workflow:
\begin{enumerate}[topsep=0pt,parsep=0pt,partopsep=0pt, leftmargin=10pt]
    \item \textbf{Producing Predicates}: These methods first generate a set of predicates \(\sP\) based on the input \(\vx\).
    \item \textbf{Generating Samples}: The underlying perturbation model \(t_{per}\) generates a set of samples \(\sX_s\) and its corresponding predicate representation set \(\sB_s\).
    \item \textbf{Learning Explanation}: The underlying learning algorithm generates the local explanation \(g_{f,\vx}\) consisting of predicates in \(\sP\) using \(\sB_s\) and \(f(\sX_s)\).
\end{enumerate}

Mainstream local model-agnostic explanation methods like Anchors, LIME, LORE, and Kernel SHAP, all follow this workflow. 
%
In the following, we introduce the main components of these explanation methods.

\textbf{Predicate Sets}\quad Given an input \(\vx\), the corresponding predicate set \(\sP\) is defined as \( \sP=\{p_i| i \in [1, d]\}\),
where \(d\) is the number of predicates in \(\sP\), a hyperparameter set by users or according to the input \(\vx\). 
Each \(p_i\) is a feature predicate that constrains the value of a set of feature values in \(\vx\), i.e. \(p_i(\vz): \bigwedge_{j\in \sA_i}1_{\vx_j=\vz_j}\), where \(\sA_i\) is the set of indices of features that \(p_i\) constrains. 
%
%
%
Then, we define the predicate representation \(\vb\in \{0,1\}^d\) of a sample input \(\vz\) as a binary vector where \(\vb_i = p_i(\vz)\). 

\textbf{Perturbation Models}\quad The perturbation model \(t_{per}\) first randomly selects \(\sB_s\subseteq \{0,1\}^{d}\) as the predicate representations of the samples.
Then, it transforms \(\sB_s\) back to the original input space to get \(\sX_s\).
For each \(\vb \in \sB_s\), 
if \(\vb_i=1\), then for each \(j\in \sA_i\), it sets \(\vz_j = \vx_j\) ; otherwise, it sets each \(\vz_j\) to a masked value, or a random value sampled from a user-defined distribution.

\textbf{Learning Algorithms}\quad Existing local model-agnostic explanation methods use different learning algorithms. Anchors uses KL-LUCB~\citep{KL-LUCB} to learn sufficient conditions, LIME and Kernel SHAP use linear regression~\citep{mcdonald2009ridge} to learn attributions, and LORE uses decision trees~\citep{YADT} to learn sufficient conditions and counterfactuals.

\textbf{Explanation Forms}\quad Existing local model-agnostic explanation methods can provide various forms of explanations for different user needs:
\begin{itemize}[topsep=0pt,parsep=0pt,partopsep=0pt, leftmargin=5pt]
    \item \textit{Attributions}: An attribution explanation~\citep{SHAP, LIME,tan2023glime, Shankaranarayana_Runje_2019, ICE,apley2020visualizing} consists of a set of feature predicates \(p_i\) with importance weights \(w_i\), i.e., \(g = \{(p_i, w_i) \mid i \in [1, d]\}\). Attributions assist users in evaluating the trustworthiness of the model~\citep{LIME}.
    \item \textit{Sufficient Conditions}: This form of explanation~\citep{Anchors} identifies a minimal set of conditions that are sufficient to produce the same output as the original input. Formally, \(f(\vz) = f(\vx)\) if \(g(\vz) = 1\), where \(g(\vz) = \bigwedge_{p \in \sQ} p(\vz)\) and \(\sQ \subseteq \sP\). Sufficient conditions help users anticipate the model's behavior on unseen inputs~\citep{Anchors}.
    \item \textit{Counterfactuals}: A counterfactual explanation~\citep{wachter2018counterfactualexplanationsopeningblack, lore} shows how a model’s prediction changes under specific input modifications. It is defined as \(f(\vz) = y\) if \(g(\vz) = 1\), where \(y \ne f(\vx)\) and \(g(\vz) = \bigwedge_{p \in \sQ} p(\vz) \wedge \bigwedge_{p \in \sC} \neg p(\vz)\), with \(\sQ, \sC \subseteq \sP\). Counterfactuals provides actionable insights for decision-making and outcome reversal~\citep{lore}.
\end{itemize}

To our knowledge, existing model-agnostic explanation methods that provide explanations other than attributions mainly provide explanations at feature levels~\citep{survey_zhang,survey_model_agnostic}.

\subsection{Concept-Based Explanations}
Concept-based explanations primarily focus on attributing high-level concepts to model predictions~\citep{concept_survey}.
While the definition of a concept may vary, \citet{molnar2020interpretable} broadly characterizes it as “an abstract idea, such as a color, an object, or even an idea,” with the key requirement that concepts be meaningful to end-users~\citep{ACE}, such as objects in images or topics in text.

Existing concept-based explanation methods have extensively explored the extraction of high-level concepts from input data, but the perturbation of high-level concepts remains largely unexplored, which is crucial for generating model-agnostic explanations.
Specifically, these approaches, including various concept bottleneck models~\citep{Koh_Nguyen_Tang_Mussmann_Pierson_Kim_Liang_2020,TBM, Oikarinen_Das_Nguyen_Weng_2023,Kim_Jung_Park_Kim_Yoon_2023}, leverage internal model representations~\citep{zhang2021invertibleconceptbasedexplanationscnn, Yeh_Kim_Arik_Li_Ravikumar_Pfister_2019, Cunningham_Ewart_Riggs_Huben_Sharkey_2023, ghorbani2019towards, Crabbé_van2022, Fel_Boutin_Béthune_Cadene_Moayeri_Andéol_Chalvidal_Serre_2023,Fel_Picard_Bethune_Boissin_Vigouroux_Colin_Cadène_Serre_2023, ACE, conceptSHAP,varshney2025generatingcounterfactualtrajectorieslatent,tapariaexplainable}, external knowledge~\citep{ConceptGlassbox, widmer2022humancompatiblexaiexplainingdata, Ciravegna_Barbiero_Giannini_Gori_Lió_Maggini_Melacci_2023}, or pre-trained models~\citep{TBM, EAC} to extract concepts from input data.
However, although \citet{Goyal_Feder_Shalit_Kim_2020} and \citet{Wu_D’Oosterlinck_Geiger_Zur_Potts_2022} try to map concept level changes back to feature level, they are designed for specific tasks, and need much effort to adapt to new datasets or models.
%
Moreover, most concept-based model-agnostic methods focus primarily on attribution-based explanations, which limits their expressiveness and applicability. Although \citet{Ciravegna_Barbiero_Giannini_Gori_Lió_Maggini_Melacci_2023}  provide rule-based sufficient conditions, they are limited to only provide global explanations, which is not tractable for complex models in practice.


\section{The \toolname Framework}
\label{sec:fram}
\label{sec:UnCLE}

\begin{figure*}[t]
    \centering
        \resizebox{0.85\textwidth}{!}{
        \includegraphics[scale=0.35]{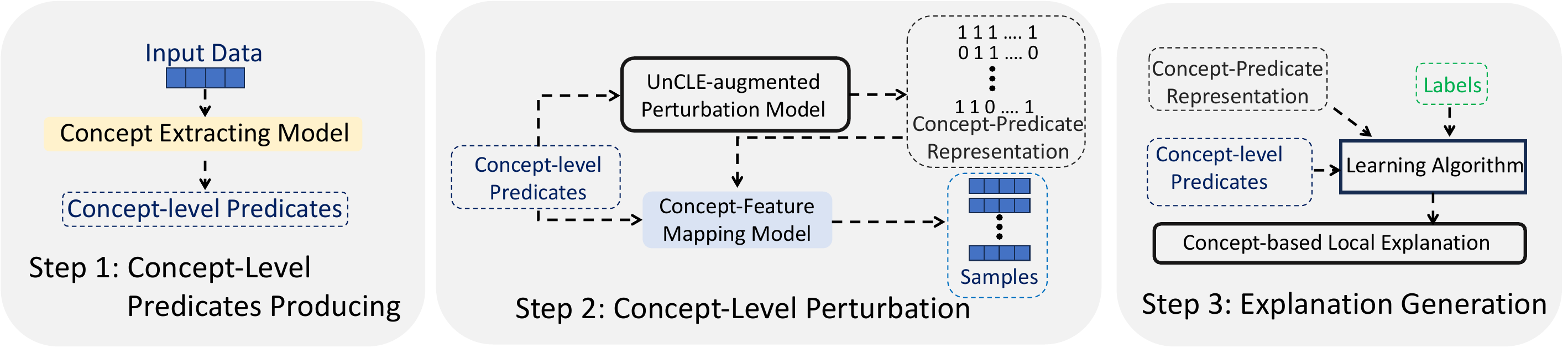} 
        }
    
\caption{The workflow of \toolname-augmented local model-agnostic explanation techniques.}
\label{fig:workflow}
\end{figure*}

In this section, we propose \toolname, a general and lightweight framework to elevate existing local model-agnostic explanation methods to concept level without significantly changing their core components. 
As shown in Figure~\ref{fig:workflow}, \toolname-augmented methods generate explanations in three steps: 1)~concept-level predicate producing, 2) concept-level perturbation, and 3) explanation generation.


\subsection{Concept-Level Predicate Producing}

We use a \textbf{concept-extracting model} to extract high-level concepts from a given input \(\vx\) based on the target task and user needs. 
As existing work has proposed several effective methods for concept extraction~\citep{ACE, ConceptGlassbox, TBM, EAC}, users can select an appropriate method based on their specific needs.

\toolname defines \textbf{concept predicates} (denoted as \(p^c\)) based on the extracted concepts.
Each concept predicate \(p^c\) is a binary function that indicates whether the input \(\vx\) satisfies a specific concept.
Subsequently, \toolname replaces the vanilla predicates set \(\sP\) with the concept predicates set \(\sP^c\). 

\textbf{An Example}\quad For the text example in Figure~\ref{fig:example-image-perturbation}(b), our approach follows \citet{TBM} to extract concepts that correlate with sentiment from a professional movie reviewer's perspective, and the model generates the four concepts. For the i-th concept, \toolname defines its corresponding predicate \(p^c_i\) as \textit{``if the sentence has the i-th concept"}, and set the predicates set \(\sP^c = \{p^c_1, p^c_2, p^c_3, p^c_4\}\).

\subsection{Concept-Level Perturbation}

The \textbf{\toolname augmented perturbation model} \(t_{per}^c\) changes high-level concepts directly to obtain samples \(\sX^c_s\) and their concept-predicate representations \(\sB^c_s\).

Specifically, \(t_{per}^c\) first generates samples in concept-predicate representations. Each \textbf{concept-predicate representation} \(\vb^c\) is a \(|\sP^c|\)-dimension binary vector corresponding to a sample \(\vz\), indicating whether the sample satisfies the concept predicates in \(\sP^c\).
Since concept predicates do not directly constrain the feature values, \(t_{per}^c\) requires a \textbf{concept-feature mapping model} \(c:~\{0,1\}^{|\sP^c|} \rightarrow \sX\) to transform the samples back to the feature level.
We find that large pre-trained models are ideal to serve as the concept-feature mapping model in \toolname, as they excel at generating content based on structured prompts.
Besides, their generated samples are more meaningful compared to simply masking to changing feature values~\citep{tan2023glime,LIME, Anchors,ReX}.
Specifically, when mapping a sample's concept-predicate representation \(\vb^c\) back to the feature level, if \(\vb^c_i = 1\), we prompt the model to ensure that the generated sample satisfies the \(i\)-th concept; if \(\vb^c_i = 0\), the generated sample must not satisfy the \(i\)-th concept.

\textbf{Examples}\quad Figure~\ref{fig:example-image-perturbation}(a) shows some samples generated by the \toolname-augmented perturbation model. For example, \(t_{per}^c\) generated the left-upper sample without a child in the image. Appendix~\ref{app:framework} shows the prompts we used to generate the samples, which does not require much effort to design.

%

\subsection{Explanation Generation}
In this stage, \toolname leverages the vanilla learning algorithms underlying existing methods to generate explanations.
%
%
Using the concept-predicate representations \(\sB^c_s\) and their corresponding outputs \(f(\sX^c_s)\), the learning algorithm learns a concept-based explanation \(g_{f,\vx}\) composed of concept predicates in \(\sP^c\). 
%
Benefiting from \toolname's generality in augmenting various existing explanation methods, \toolname inherits their ability to provide various forms of explanations, including attributions, sufficient conditions, and counterfactuals.

\toolname enables the generation of multiple explanation forms through a unified framework with a single click.
As a result, users can flexibly choose the explanation type that best fits their needs.
We refer to the resulting collection of explanations as a \textbf{\toolname local unified explanation}.

\textbf{Examples}\ \ \ Figure~\ref{fig:example-image-perturbation},~\ref{fig:example-image1} show examples of \toolname generated attributions, sufficient conditions, and counterfactuals.

\section{Empirical Evaluation}
\label{sec:eval}

We evaluate \toolname from three aspects: (1) the fidelity of concept-level perturbation, (2) the fidelity of \toolname-augmented explanations, and (3) how much \toolname helps users use explanations in downstream tasks. 
We also discuss the time efficiency of \toolname and the robustness of \toolname to different generative model choices.

\subsection{Perturbation Fidelity Evaluation}
\label{sec:eval-pert}

To ensure \toolname can faithfully generate explanations, it is essential to ensure the underlying large models perform accurate concept-feature mapping.

\textbf{Experimental Setup}\quad
We conducted experiments on two types of input data: text and image. 

For text data, we use DeepSeek-V3~\citep{liu2024deepseekv3} as the concept–feature mapping model.
We verify its fidelity on perturbing on both human-annotated and model-extracted concepts.
We use four datasets and randomly select 250 sentences from each of the datasets.
For each sentence, we instruct DeepSeek-V3 to generate 10 perturbed samples by altering specific given concepts, and using a checker to verify whether the generated sentences meet the requirements.

Specifically, for human-annotated concepts, we use two named entity recognition datasets, 
CoNLL-2003~\citep{tjong-kim-sang-de-meulder-2003-introduction} and OntoNotes 5.0~\citep{pradhan2013towards}. The concepts include entity types such as \textit{person}, \textit{location}, and \textit{organization}.
We use a fine-tuned BERT model~\citep{devlin2018bert} as the checker.
%
%
For model-extracted concepts, as \citet{TBM} have verified that GPT-based models can extract high-quality textual concepts and verify concept satisfaction, we use GPT-4o to extract concepts from sentences of the Large Movie Review~\citep{IMDB} and Fake News~\citep{fake-news-dataset} datasets, and also use GPT-4o as the checker.
%


For image data, we used Blended Latent Diffusion~\citep{LDM} as the concept-feature mapping model, and conducted our experiment on images from the validation set of
COCO dataset~\citep{COCO}, where objects in images are annotated by humans. For each image, we generated 10 samples by altering these concepts, and checked if the generated samples satisfy the requirements by an object-detection YOLO11 model~\citep{YOLO11}.

\textbf{Metrics and Results}\quad
We evaluated perturbation fidelity using accuracy, i.e. the proportion of generated samples that satisfy the given concept requirements. 
Table~\ref{table:perutrbation} shows that our perturbation models achieve an average accuracy of 96.8\%, indicating that \toolname can faithfully generate samples that meet the given concept requirements.


\subsection{Explanation Fidelity Evaluation}
\label{sec:eval-exp}
Our evaluation consists of two parts: (1) assessing the fidelity improvement that \toolname brings to existing local model-agnostic explanation methods (Anchors, LIME, LORE, and Kernel SHAP (KSHAP for short)), and (2) comparing the fidelity of \toolname with state-of-the-art concept-based methods: TBM~\citep{TBM} and LACOAT~\citep{yu2024latent} for text models, and EAC~\citep{EAC} and ConceptLIME~\citep{tan2024concept} for image models.
%


\begin{table}
    \caption{Accuracy of using different concept-level perturbation models to generate samples that satisfy certain concept predicates. We used DeepSeekV3 and Blended Latent Diffusion in our perturbation and explanation fidelity evaluation in Section~\ref{sec:eval-pert} and~\ref{sec:eval-exp}.}
    \centering
    {
    \small
    \resizebox{\linewidth}{!}{
     \begin{tabular}{@{}lcccccc@{}}
        \toprule
        Model & CoNLL & Onto. & Large. & Fake. & COCO \\
        \midrule
        \rowcolor{gray!10}\textbf{DeepSeekV3} & 97.5 & 98.1& 97.7 & 94.7 & -- \\
        Qwen2.5 72B & 98.1 & 97.3& 97.5 & 95.0 & -- \\
        Qwen2.5 7B & 92.4 & 91.9& 92.3 & 91.3 & --\\[0.5em]
        \rowcolor{gray!10}\textbf{Blended Latent Diffusion} & -- & -- & -- & -- & 98.1 \\
        Latent Consistency Model & -- & -- & -- & -- & 97.5 \\
        \bottomrule
    \end{tabular}
    }
    }
    \label{table:perutrbation}
\end{table}

\begin{table*}[t]
    \caption{Average coverage and precision (higher is better) of Anchors, LORE, and their \toolname-augmented versions.}
    \centering
    \setlength{\tabcolsep}{1mm}
    \resizebox{0.7\textwidth}{!}{
    \small
    \begin{tabular}{@{}lcccccccc@{}}
        \toprule
        \multirow{2}{*}{Models} & \multicolumn{4}{c}{Coverage (\%) $\uparrow$} & \multicolumn{4}{c}{Precision (\%) $\uparrow$} \\
        \cmidrule(lr){2-5} \cmidrule(lr){6-9}
                                & Anchors & Anchors* & LORE & LORE* & Anchors & Anchors* & LORE & LORE* \\
        \midrule
        DeepSeek-V3/Large.      & $4.7\pm0.7$ & $\mathbf{23.5}\pm 2.0$ & $2.3 \pm 0.3$ & $\mathbf{22.4} \pm 2.1$ & $79.2 \pm 1.3$ & $\mathbf{96.2}\pm 0.1$ & $63.1\pm 1.7$ & $\mathbf{73.9}\pm 2.2$ \\
        DeepSeek-V3/Fake.       & $3.4\pm0.7$ & $\mathbf{22.8}\pm1.7$ & $2.4\pm0.5$ & $\mathbf{18.5}\pm2.0$ & $80.5\pm0.3$ & $\mathbf{93.6}\pm0.5$ & $60.4\pm2.5$ & $\mathbf{73.1} \pm 2.4$ \\[0.2em]
        BERT/Large.             & $3.9 \pm 0.4$ & $\mathbf{24.5} \pm 1.4$ & $3.6 \pm 0.5$ & $\mathbf{20.3} \pm 1.8$ & $81.7 \pm 0.2$ & $\mathbf{91.0} \pm 0.8$ & $64.2 \pm 1.1$ & $\mathbf{79.4} \pm 1.1$ \\
        BERT/Fake.              & $4.1 \pm 0.3$ & $\mathbf{24.2} \pm 1.7$ & $2.9 \pm 0.1$ & $\mathbf{22.6} \pm 1.7$ & $78.2 \pm 1.3$ & $\mathbf{89.8} \pm 0.8$ & $62.2 \pm 2.5$ & $\mathbf{76.5} \pm 2.1$ \\[0.4em]
        YOLOv8/ImageNet         & $23.5 \pm 1.5$ & $\mathbf{31.9} \pm 1.7$ & $19.4 \pm 1.8$ & $\mathbf{26.1} \pm 2.1$ & $78.3 \pm 3.2$ & $\mathbf{95.2} \pm 1.3$ & $68.3 \pm 2.3$ & $\mathbf{81.8} \pm 2.5$ \\
        YOLOv8/Caltech101       & $16.2 \pm 1.3$ & $\mathbf{23.6} \pm 1.6$ & $18.7 \pm 1.3$ & $\mathbf{23.5} \pm 1.5$ & $84.3 \pm 0.5$ & $\mathbf{96.7} \pm 0.8$ & $72.9 \pm 1.3$ & $\mathbf{82.1} \pm 2.1$ \\
        YOLOv8/CUB              & $17.8 \pm1.5$ & $\mathbf{22.5}\pm 1.1$ & $20.1\pm 1.2$ & $\mathbf{24.1}\pm 1.6$ & $83.5\pm 0.1$ & $\mathbf{95.5}\pm0.2$ & $68.8\pm2.7$ & $\mathbf{75.8}\pm1.7$ \\[0.2em]
        ViT/ImageNet            & $22.3 \pm 1.2$ & $\mathbf{31.5} \pm 1.4$ & $20.7 \pm 1.1$ & $\mathbf{23.9} \pm 1.5$ & $82.4 \pm 0.5$ & $\mathbf{93.6} \pm 0.3$ & $75.2 \pm 0.9$ & $\mathbf{84.3} \pm 1.2$ \\
        ViT/Caltech101          & $18.9 \pm 1.8$ & $\mathbf{26.7} \pm 2.0$ & $19.4 \pm 1.7$ & $\mathbf{24.1} \pm 1.4$ & $83.5 \pm 0.8$ & $\mathbf{97.0} \pm 0.4$ & $71.2 \pm 1.2$ & $\mathbf{80.9} \pm 0.7$ \\
        ViT/CUB                 & $21.9\pm1.0$ & $\mathbf{29.1}\pm1.0$ & $26.4 \pm 1.1$ & $\mathbf{30.7} \pm 1.0$ & $85.6\pm0.3$ & $\mathbf{94.3}\pm1.2$ & $74.5 \pm 1.3$ & $\mathbf{82.0} \pm 1.0$ \\[0.2em]
        ResNet-50/ImageNet      & $21.5 \pm 1.2$ & $\mathbf{29.9} \pm 1.6$ & $20.1 \pm 1.4$ & $\mathbf{29.3} \pm 2.1$ & $81.8 \pm 2.3$ & $\mathbf{97.9} \pm 0.5$ & $76.8 \pm 2.1$ & $\mathbf{83.6} \pm 1.8$ \\
        ResNet-50/Caltech101    & $23.4 \pm 1.2$ & $\mathbf{27.8} \pm 1.2$ & $19.7 \pm 1.3$ & $\mathbf{28.4} \pm 1.2$ & $84.5 \pm 4.4$ & $\mathbf{96.4} \pm 1.0$ & $77.3 \pm 1.4$ & $\mathbf{85.1} \pm 1.0$ \\
        ResNet-50/CUB           & $17.0\pm 1.1$ & $\mathbf{26.1}\pm 1.2$ & $19.1\pm1.4$ & $\mathbf{20.6}\pm1.3$ & $78.3\pm 4.3$ & $\mathbf{89.5} \pm 0.2$ & $56.0\pm2.0$ & $\mathbf{72.7}\pm3.9$ \\[0.4em]
        Qwen2.5-VL              & $10.5 \pm 0.9$ & $\mathbf{21.8} \pm 1.9$ & $8.5 \pm 1.1$ & $\mathbf{21.8} \pm 1.9$ & $75.4 \pm 2.3$ & $\mathbf{93.1} \pm 0.5$ & $64.7 \pm 1.2$ & $\mathbf{73.2} \pm 1.4$ \\
        \bottomrule
    \end{tabular}
    }
    \label{table:coverage_precision}
\end{table*}

\begin{table*}[t]
    \caption{Average AOPC (higher is better) and $\mathrm{accuracy_a}$ (lower is better) of LIME, KSHAP, and their \toolname-augmented versions.}
    \centering
    \setlength{\tabcolsep}{1mm}
    \resizebox{0.7\textwidth}{!}{
    \small
    \begin{tabular}{@{}lcccccccc@{}}
        \toprule
        \multirow{2}{*}{Models} & \multicolumn{4}{c}{AOPC (\%)$\uparrow$} & \multicolumn{4}{c}{$\mathrm{Accuracy_a}$ (\%) $\downarrow$} \\
        \cmidrule(lr){2-5} \cmidrule(lr){6-9}
                                & LIME & LIME* & KSHAP & KSHAP* & LIME & LIME* & KSHAP & KSHAP* \\
        \midrule
        DeepSeek-V3/Large.      & $22.1 \pm 4.1$ & $\mathbf{46.4} \pm 3.3$ & $32.7 \pm 3.3$ & $\mathbf{52.5} \pm 3.3$ & $77.3 \pm 1.9$ & $\mathbf{49.3} \pm 0.8$ & $73.1 \pm 1.7$ & $\mathbf{46.3} \pm 1.2$ \\
        DeepSeek-V3/Fake.       & $21.7 \pm 1.7$ & $\mathbf{48.3} \pm 2.1$ & $30.4 \pm 1.4$ & $\mathbf{49.6} \pm 1.9$ & $76.4\pm1.3$ & $\mathbf{51.1}\pm 1.1$ & $71.6\pm1.4$ & $\mathbf{49.6}\pm1.1$ \\[0.2em]
        BERT/Large.             & $24.3 \pm 3.2$ & $\mathbf{45.6} \pm 3.6$ & $37.9 \pm 4.1$ & $\mathbf{55.3} \pm 2.3$ & $75.7 \pm 1.2$ & $\mathbf{47.7} \pm 1.9$ & $60.3 \pm 1.7$ & $\mathbf{40.2} \pm 2.1$ \\
        BERT/Fake.              & $25.7 \pm 2.7$ & $\mathbf{47.2} \pm 3.1$ & $33.8 \pm 3.0$ & $\mathbf{51.9} \pm 3.4$ & $76.2 \pm 1.3$ & $\mathbf{53.2} \pm 2.0$ & $67.9 \pm 1.2$ & $\mathbf{45.1} \pm 1.5$ \\[0.4em]
        YOLOv8/ImageNet         & $38.9 \pm 2.0$ & $\mathbf{51.4} \pm 1.8$ & $41.7 \pm 1.5$ & $\mathbf{58.7} \pm 1.7$ & $14.9 \pm 1.1$ & $\mathbf{4.7} \pm 1.3$ & $23.9 \pm 1.3$ & $\mathbf{6.5} \pm 0.9$ \\
        YOLOv8/Caltech101       & $44.0 \pm 4.0$ & $\mathbf{62.2} \pm 2.8$ & $43.1 \pm 3.3$ & $\mathbf{59.6} \pm 3.6$ & $23.8 \pm 1.5$ & $\mathbf{13.7} \pm 1.2$ & $21.1 \pm 1.6$ & $\mathbf{11.3} \pm 1.6$ \\
        YOLOv8/CUB              & $51.8 \pm 3.2$ & $\mathbf{63.7} \pm 3.0$ & $52.9 \pm 3.1$ & $\mathbf{62.4} \pm 3.2$ & $7.0\pm1.2$ & $\mathbf{1.0}\pm0.5$ & $8.1\pm1.4$ & $\mathbf{3.2} \pm 1.2$ \\[0.2em]
        ViT/ImageNet            & $45.8 \pm 2.8$ & $\mathbf{56.3} \pm 1.7$ & $46.8 \pm 2.1$ & $\mathbf{62.0} \pm 3.3$ & $22.4 \pm 1.2$ & $\mathbf{5.4} \pm 1.0$ & $18.6 \pm 1.4$ & $\mathbf{5.3} \pm 1.4$ \\
        ViT/Caltech101          & $47.3 \pm 2.8$ & $\mathbf{64.5} \pm 3.5$ & $48.1 \pm 3.0$ & $\mathbf{67.3} \pm 3.4$ & $22.3 \pm 1.3$ & $\mathbf{13.4} \pm 1.2$ & $21.3 \pm 1.9$ & $\mathbf{11.8} \pm 1.5$ \\
        ViT/CUB                 & $59.2 \pm 2.1$ & $\mathbf{70.8} \pm 2.1$ & $62.9 \pm 2.2$ & $\mathbf{69.7} \pm 2.2$ & $6.5\pm0.8$ & $\mathbf{1.6}\pm0.3$ & $8.2\pm1.0$ & $\mathbf{3.2}\pm0.7$ \\[0.2em]
        ResNet-50/ImageNet      & $22.3 \pm 3.0$ & $\mathbf{32.3} \pm 2.9$ & $24.1 \pm 2.8$ & $\mathbf{33.9} \pm 2.4$ & $21.1  \pm 1.4$ & $\mathbf{3.6} \pm 1.0$ & $19.4 \pm 0.8$ & $\mathbf{3.9} \pm 0.9$ \\
        ResNet-50/Caltech101    & $29.7 \pm 3.7$ & $\mathbf{48.1} \pm 3.2$ & $27.5 \pm 3.4$ & $\mathbf{50.1} \pm 2.8$ & $17.7 \pm 1.3$ & $\mathbf{7.9} \pm 1.2$ & $15.8 \pm 1.3$ & $\mathbf{6.3} \pm 1.0$ \\
        ResNet-50/CUB           & $46.5 \pm 3.0$ & $\mathbf{56.3} \pm 3.5$ & $45.2 \pm 3.4$ & $\mathbf{56.7} \pm 3.0$ & $4.9 \pm 1.1$ & $\mathbf{2.1} \pm 0.9$ & $7.4\pm1.6$ & $\mathbf{4.5}\pm1.0$ \\[0.4em]
        Qwen2.5-VL              & $24.5 \pm 2.1$ & $\mathbf{40.7} \pm 2.4$ & $31.6 \pm 2.8$ & $\mathbf{39.8} \pm 2.7$ & $21.2 \pm1.2$ & $\mathbf{5.1} \pm 0.9$ & $16.9 \pm 1.4$ & $\mathbf{3.0} \pm 0.6$ \\
        \bottomrule
    \end{tabular}
    }
    \label{table:aopc_accuracy}
\end{table*}

\subsubsection{Experimental Setup}
\label{sec:evelset}
We conducted the experiments on two text classification tasks, three image classification tasks, and one multimodal task.

\textbf{Text tasks}\quad Text classification models take a text sequence as input and predict its category.
%
%
We used Large Movie Review~\citep{IMDB} and Fake News~\citep{fake-news-dataset} datasets, and predicted the labels of sentences in the validation set using a fine-tuned BERT~\citep{devlin2018bert} and DeepSeek-V3 with in-context examples, which are the models to explain.
For the baseline explanation methods, we kept the default settings.
For \toolname, we followed the settings in the perturbation fidelity evaluation.
%
%

\textbf{Image tasks}\quad Image classification models take an image as input and predict its category.
%
We conducted experiments on ImageNet~\citep{deng2009imagenet}, Caltech-101~\citep{caltech101}, and CUB-200~\citep{welinder2010caltech}. 
For each dataset, we used a fine-tuned YOLOv8, Vision Transformer (ViT)~\citep{oquab2023dinov2,darcet2023vitneedreg}, and ResNet-50~\citep{he2016deep} to classify images from the validation set and explain the models' local behaviors.
For the vanilla methods, we followed LIME to use Quickshift~\citep{jiang2018quickshiftprovablygoodinitializations} to obtain superpixels from input images as predicates.
For \toolname, we used SAM~\citep{kirillov2023segany} as the concept-extracting model, and Blended Latent Diffusion as the concept-feature mapping model.
For EAC and ConceptLIME, we kept their default settings.

\textbf{Multimodal tasks}\quad We follow the settings in text and image tasks, explain multimodal Qwen2.5-VL on 250 randomly selected Yes/No Question of the VQAv2~\citep{Goyal2017Making} dataset, and compare the fidelity of \toolname-augmented explanations with the vanilla ones.

As TBM, LACOAT, EAC, and ConceptLIME are not applicable to multimodal tasks, we compare the fidelity of these methods and \toolname-augmented explanations only on text and image tasks.



\subsubsection{Evaluation Metrics}
Fidelity reflects how faithfully an explanation describes a target model. 
As the explanation form of these methods varies, we used different metrics to evaluate their fidelity.

For Anchors and LORE, following the settings of their original papers, we used \textbf{coverage} and \textbf{precision} as the fidelity metrics (which are named differently in the LORE paper). Given a target model $f$, an input $\vx$, and a distribution $D_\vx$ derived from the perturbation model, and the corresponding explanation $g$, we defined the coverage as \(
\cov(\vx;f,g) = \E_{\vz\sim D_\vx}[g(\vz)]\), which indicates the proportion of inputs in the distribution that match the rule.
We defined precision as \(\prec(\vx;f,g) = \E_{\vz\sim D_\vx}[\1_{f(\vz)=y} | g(\vz)]\), where $y$ is the consequence of the rules in $g$ with $y = f(\vx)$ for factual rules and $y\neq f(\vx)$ for counterfactual rules.
Precision indicates the proportion of covered inputs for which $g$ correctly predicts the model outputs. 

As LIME and KSHAP are attribution-based local surrogates, we measured their explanation fidelity by \textbf{deletion experiments}. We used \textit{Area Over most relevant first perturbation curve} (AOPC), and $\mathrm{accuracy_a}$ as the metrics~\citep{samek2016evaluating, DecompX}.
Given a target model $f$, an input $\vx$, the model output $y = f(\vx)$, their corresponding explanation $g$, and $\vx^{(k)}$ that is generated by masking the $k\%$ most important predicates in $\vx$,
AOPC and $\mathrm{accuracy_a}$ are defined as follows:
\begin{itemize}[noitemsep,topsep=0pt,parsep=0pt,partopsep=0pt, leftmargin=5pt]
    \item \textbf{AOPC:}  Let \(
        \AOPC_k = \frac{1}{|\sT|} \sum_{\vx \in \sT} (p_f(y|\vx)-p_f(y|\vx^{(k)})), 
    \)
    where $p_f(y|\vx)$ is the probability of $f$ to output $y$ given the input $\vx$, and $\sT$ is the set of all test inputs.
    $\AOPC_k$ indicates the average change of the model output when masking the $k\%$ most important predicates. A higher $\AOPC_k$ indicates a better explanation. AOPC is defined as $\AOPC = \sum_{k=1}^{100} \AOPC_k/100$.

    \item \textbf{Accuracy\textsubscript{a}}: $\mathrm{Accuracy_a}$ indicates the proportion of inputs among all $\vx^{(k)}$ that the target model gives the same output as the original input $\vx$, i.e. $\E[f(\vx^{(k)}) = f(\vx)]$. Specifically, $\mathrm{accuracy_a}$ is different from the standard accuracy, and a lower $\mathrm{accuracy_a}$ indicates a better explanation.

\end{itemize}
Specifically, we only considered the predicates that positively contribute to $f(\vx)$.

When comparing to state-of-the-art concept-based explanation methods, we considered \toolname unified explanations as one of the main advantages of \toolname over existing concept-based approaches is offering rich forms of explanations.
Since there are multiple forms of explanations, considering that they can all serve as local surrogate models, we defined the metrics as follows~\citep{balagopalan2022road,ismail2021improving}: given a target model $f$, an input $\vx$, a perturbation distribution $\sD_{\vx}$, their corresponding explanation $g$, and a performance metric $L$ (e.g. accuracy, F1 score, etc.), we defined the (in-)fidelity as \(E_{\vz \sim \sD_\vx}[L(\mathbf{1}_{\{\vz|f(\vz)=f(\vx)\}}(\vz), g(\vz))]\), which indicates the performance of using $g$ to approximate the local behavior of $f$, i.e., predict if f($\vz$) is the same as $f(\vx)$.   
Here, we used accuracy as the performance metric.


%
For a fair comparison, we calculate fidelity in different neighborhoods for each setup:
1) When comparing \toolname-augmented explanations with their feature-level versions, we use feature-level neighborhood.
2) When comparing \toolname unified explanations with other concept-based explanations, we use concept-level neighborhood.

%
%
%
%

\subsubsection{Results}
Table~\ref{table:coverage_precision} and~\ref{table:aopc_accuracy} show the fidelity of Anchors, LORE, LIME, KSHAP, and their \toolname-augmented versions.
For Anchors and LORE, \toolname improves the average coverage by 11.2\% and 9.5\%, and the average precision by  13.0\% and 10.6\%, respectively.
For LIME and KSHAP, \toolname improves the average AOPC by 0.164 and 0.151, and decreases the 
average $\mathrm{accuracy_a}$ by 14.8\% and 13.8\%, respectively.
%
%
%
%
We did paired t-tests for setups that only differ in whether to apply \toolname, which indicates with over 99\% confidence the improvement is significant. 

Table~\ref{table:unified} shows the fidelity of \toolname-augmented explanations and state-of-the-art concept-based explanation methods: TBM and LACOAT for text tasks, and EAC and ConceptLIME for image tasks, which shows that \toolname helps KSHAP to achieve higher fidelity than these methods on all tasks, and \toolname local unified explanations further achieve 4.52\% more fidelity than \toolname-augmented KSHAP explanations.
%

%


\begin{table}[t]
    \centering
    \caption{Average accuracy (\%) (higher accuracy is better) of TBM, EAC, \toolname-augmented KSHAP (denoted as KSHAP*), and \toolname unified explanations. }
    \setlength{\tabcolsep}{1mm}
    \small
    \resizebox{\linewidth}{!}{
     \begin{tabular}{@{}lcccccc@{}}
        \toprule
        Models & TBM & LACOAT & EAC & ConceptLIME & KSHAP* & \toolname \\
        \midrule
        DeepSeek-V3/Large. & 78.3 & 65.8 & -- & -- & 87.1 & \textbf{90.7} \\
        DeepSeek-V3/Fake. & 75.4 & 66.7 & -- & -- & 82.3 & \textbf{87.3}\\
        BERT/Large. & 82.4 & 73.4 & -- & -- & 86.3 & \textbf{92.6}\\
        BERT/Fake. & 81.4 & 75.2 & -- & -- & 85.6 & \textbf{91.8} \\[0.2em]

        YOLOv8/ImageNet & -- & -- & 82.7 & 81.3 & 88.1 & \textbf{92.4} \\
        YOLOv8/Caltech101 & -- & -- & 84.6 & 83.9 & 88.7 & \textbf{92.1} \\
         YOLOv8/CUB & -- & -- & 76.8 &72.6 & 86.7 &\textbf{89.3}\\
        ViT/ImageNet & -- & -- & 83.4 & 79.3 & 88.6 & \textbf{91.6} \\
        ViT/Caltech101 & -- & -- & 82.7 & 84.1 & 89.4 & \textbf{92.6} \\
        ViT/CUB & -- & -- & 74.3 & 72.1 & 86.4 & \textbf{90.3}\\
        ResNet-50/ImageNet & -- & -- & 78.1 & 77.9 & 85.0 & \textbf{90.2} \\
        ResNet-50/Caltech101 & -- & -- & 79.7 & 76.8 & 85.4 & \textbf{91.5} \\
        ResNet-50/CUB & -- & -- & 77.4 & 77.1 & 83.1 & \textbf{89.1}\\
        \bottomrule
    \end{tabular}
    }
    \label{table:unified}
\end{table}


\subsection{Human Evaluation}

\toolname provides explanations in various forms beyond attributions, including sufficient conditions and counterfactuals, which can more effectively support users in downstream decision-making tasks.

Specifically, sufficient conditions help users anticipate the model’s behavior on unseen inputs, while counterfactuals help understand how the model’s prediction changes under specific input modifications.
To evaluate the usefulness of introducing these explanation forms, we conducted a user study following the setup in~\citet{Anchors, warren2022featuresexplainabilityusersunderstand, hase-bansal-2020-evaluating}, comparing our concept-based sufficient conditions and counterfactuals with concept-based attribution explanations provided by EAC~\citep{EAC} across two tasks.

We recruited 18 subjects, all graduate students with machine learning experience but no explainable AI expertise. Each subject completed a questionnaire with 10 tests per task.

For the sufficient condition task, in each test, participants were first shown an image randomly sampled from the ImageNet dataset, along with YOLOv8's prediction and an explanation randomly selected from EAC or \toolname-augmented Anchors explanations.
%
They were then shown 5 new images generated by concept-level perturbing the original image, and asked to predict whether the model would output the same label for each new image based on the explanation. They could answer “yes” or “no.”
We evaluate responses using two metrics: \(\mathrm{coverage}_u\), the proportion of “yes” responses, and \(\mathrm{precision}_u\), the proportion of correct predictions among the “yes” responses.

The counterfactual task follows a similar setup, except that participants were shown \toolname-augmented LORE explanations instead of Anchors and were asked to predict whether the model would output a \emph{different} label for each new image.


Table~\ref{table:userstudy} shows the average \(\mathrm{coverage}_u\) and \(\mathrm{precision}_u\) of the two tasks.
The results show that \toolname-augmented explanations outperform EAC by 3.0\% in \(\mathrm{coverage}_u\) and 8.1\% in \(\mathrm{precision}_u\) for the sufficient condition task, and by 6.8\% and 14.2\% respectively for the counterfactual task.
These findings indicate that \toolname-augmented explanations are more effective in helping users apply explanations to reason about model behavior.



\begin{table}
    \caption{Human Evaluation Results on EAC, \toolname-augmented Anchors (denoted as Anchors*), and \toolname-augmented LORE (denoted as LORE*).}
    \centering
    \resizebox{0.7\linewidth}{!}{
    \small
    \begin{tabular}{@{}lcccc@{}}
        \toprule
        \multirow{2}{*}{Methods} & \multicolumn{2}{c}{Sufficient Conditions} & \multicolumn{2}{c}{Counterfactuals} \\
        \cmidrule(lr){2-3} \cmidrule(lr){4-5}
        & EAC & Anchors* & EAC & LORE* \\
        \midrule
        \(\mathrm{coverage}_u\) (\%)  & 57.0 & 60.0 & 29.3 & 36.1\\
         \(\mathrm{Precision}_u\) (\%) & 63.9 & 72.0 & 82.8 & 97.0\\
        \bottomrule
    \end{tabular}
    }
    \label{table:userstudy}
\end{table}

\subsection{Runtime Overhead}
\label{sec:eval-overhead}

\begin{figure}[t]
    \centering
    \resizebox{\linewidth}{!}{
        \includegraphics{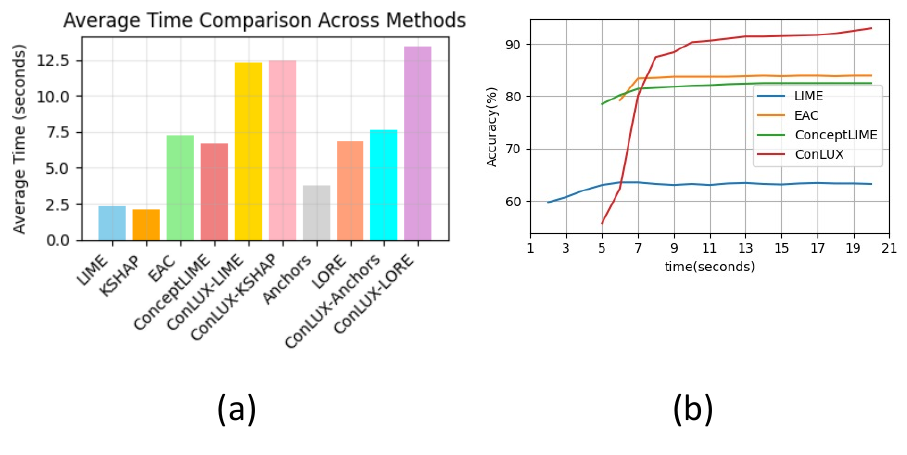}
    }
    \caption{(a) Average running time of explanation methods we used on image data. (b) Matched-budget comparisons between LIME, EAC, ConceptLIME and \toolname local unified explanations.}
    \label{fig:time}
\end{figure}

As \toolname relies on large pre-trained models to perform concept–feature mapping, it inevitably introduces additional computational overhead compared to vanilla explanation methods.
We show the execution time of the explanation fidelity experiments on image data in Figure~\ref{fig:time}.
As shown in Figure~\ref{fig:time}(a), \toolname does require additional computation time due to generative model calls. However, the running time of \toolname-augmented methods is practically acceptable.

We further conducted matched-budget comparisons (measured in running time) to evaluate the utility of each method under the same computational constraints. 
We followed the experimental settings in Section~\ref{sec:evelset}, explaining ViT on the ImageNet dataset with LIME, EAC, and ConceptLIME, and \toolname local unified explanations. Figure~\ref{fig:time}(b) shows the results.
While \toolname may start with lower fidelity at very low budgets due to the overhead of generative model calls, it surpasses baseline methods as the budget increases. 
This indicates that when users have sufficient computational time, \toolname can effectively utilize it to provide higher-fidelity explanations. Especially when \toolname surpasses baseline methods, the absolute time cost is still not high.

\subsection{Robustness to Concept-Feature Mapping Models}
\label{sec:eval-robust}
As \toolname relies on generative models for concept–feature mapping, we evaluate its robustness under different generative model choices.
We replace the generative model used in \toolname with alternative ones and measure the resulting perturbation fidelity.
Specifically, we test two additional Qwen2.5 models for text perturbations and Latent Consistency Model~\citep{LCM} for image perturbations.
Table~\ref{table:perutrbation} summarizes the results, showing that \toolname maintains high perturbation fidelity across all tested models.

This demonstrates that \toolname is flexible in choosing the generative model for concept–feature mapping, which brings two benefits: (1) users can select models based on their own preferences or resource constraints, and (2) users can employ multiple generative models to increase the diversity of generated samples and reduce potential biases from relying on a single model.

Further results and discussion are provided in Appendix~\ref{app:evelset}.
\section{Conclusion}
\label{sec:conclusion}
We have proposed \toolname, a general framework that elevates local model-agnostic explanation methods to the concept level.
%
\toolname offers unified explanations combining attributions, sufficient conditions, and counterfactuals. This satisfies diverse user needs and fills the current gap of concept-based explanations lacking forms beyond attributions.
\toolname achieves this by utilizing large pre-trained models to extract high-level concepts, and extending perturbation models to sample in the concept space.
We have instantiated \toolname on Anchors, LIME, LORE, and Kernel SHAP, and demonstrated the state-of-the-art performance of \toolname by empirical evaluations.
\toolname shows that it is unnecessary to design concept-based explanation methods from scratch, as existing local model-agnostic methods can be easily elevated to concept level in a lightweight manner.

\section*{Impact Statement}

This paper presents work whose goal is to advance the field of Machine Learning. There are many potential societal consequences of our work, none of which we feel must be specifically highlighted here.

\bibliography{conlux}

@inproceedings{Anchors,
  author    = {Marco T{\'{u}}lio Ribeiro and
               Sameer Singh and
               Carlos Guestrin},
  editor    = {Sheila A. McIlraith and
               Kilian Q. Weinberger},
  title     = {Anchors: High-Precision Model-Agnostic Explanations},
  booktitle = {Proceedings of the Thirty-Second {AAAI} Conference on Artificial Intelligence,
               (AAAI-18), the 30th innovative Applications of Artificial Intelligence
               (IAAI-18), and the 8th {AAAI} Symposium on Educational Advances in
               Artificial Intelligence (EAAI-18), New Orleans, Louisiana, USA, February
               2-7, 2018},
  pages     = {1527--1535},
  publisher = {{AAAI} Press},
  year      = {2018}
}

@article{apley2020visualizing,
  title     = {Visualizing the effects of predictor variables in black box supervised learning models},
  author    = {Apley, Daniel W and Zhu, Jingyu},
  journal   = {Journal of the Royal Statistical Society Series B: Statistical Methodology},
  volume    = {82},
  number    = {4},
  pages     = {1059--1086},
  year      = {2020},
  publisher = {Oxford University Press}
}

@inproceedings{balagopalan2022road,
  title     = {The road to explainability is paved with bias: Measuring the fairness of explanations},
  author    = {Balagopalan, Aparna and Zhang, Haoran and Hamidieh, Kimia and Hartvigsen, Thomas and Rudzicz, Frank and Ghassemi, Marzyeh},
  booktitle = {Proceedings of the 2022 ACM Conference on Fairness, Accountability, and Transparency},
  pages     = {1194--1206},
  year      = {2022}
}

@article{concept_survey,
  title        = {Concept-based Explainable Artificial Intelligence: A Survey},
  url          = {http://arxiv.org/abs/2312.12936},
  abstractnote = {The field of explainable artificial intelligence emerged in response to the growing need for more transparent and reliable models. However, using raw features to provide explanations has been disputed in several works lately, advocating for more user-understandable explanations. To address this issue, a wide range of papers proposing Concept-based eXplainable Artificial Intelligence (C-XAI) methods have arisen in recent years. Nevertheless, a unified categorization and precise field definition are still missing. This paper fills the gap by offering a thorough review of C-XAI approaches. We define and identify different concepts and explanation types. We provide a taxonomy identifying nine categories and propose guidelines for selecting a suitable category based on the development context. Additionally, we report common evaluation strategies including metrics, human evaluations and dataset employed, aiming to assist the development of future methods. We believe this survey will serve researchers, practitioners, and domain experts in comprehending and advancing this innovative field.},
  note         = {arXiv:2312.12936 [cs]},
  number       = {arXiv:2312.12936},
  publisher    = {arXiv},
  author       = {Poeta, Eleonora and Ciravegna, Gabriele and Pastor, Eliana and Cerquitelli, Tania and Baralis, Elena},
  year         = {2023},
  month        = dec,
  language     = {en}
}

@article{ConceptGlassbox,
  title        = {ConceptGlassbox: Guided Concept-Based Explanation for Deep Neural Networks},
  volume       = {16},
  issn         = {1866-9964},
  doi          = {10.1007/s12559-024-10262-8},
  abstractnote = {Various industries and fields have utilized machine learning models, particularly those that demand a significant degree of accountability and transparency. With the introduction of the General Data Protection Regulation (GDPR), it has become imperative for machine learning model predictions to be both plausible and verifiable. One approach to explaining these predictions involves assigning an importance score to each input element. Another category aims to quantify the importance of human-understandable concepts to explain global and local model behaviours. The way concepts are constructed in such concept-based explanation techniques lacks inherent interpretability. Additionally, the magnitude and diversity of the discovered concepts make it difficult for machine learning practitioners to comprehend and make sense of the concept space. To this end, we introduce ConceptGlassbox, a novel local explanation framework that seeks to learn high-level transparent concept definitions. Our approach leverages human knowledge and feedback to facilitate the acquisition of concepts with minimal human labelling effort. The ConceptGlassbox learns concepts consistent with the user’s understanding of a concept’s meaning. It then dissects the evidence for the prediction by identifying the key concepts the black-box model uses to arrive at its decision regarding the instance being explained. Additionally, ConceptGlassbox produces counterfactual explanations, proposing the smallest changes to the instance’s concept-based explanation that would result in a counterfactual decision as specified by the user. Our systematic experiments confirm that ConceptGlassbox successfully discovers relevant and comprehensible concepts that are important for neural network predictions.},
  number       = {5},
  journal      = {Cognitive Computation},
  author       = {El Shawi, Radwa},
  year         = {2024},
  month        = sep,
  pages        = {2660–2673},
  language     = {en}
}

@article{conceptSHAP,
  title   = {On completeness-aware concept-based explanations in deep neural networks},
  volume  = {33},
  journal = {Advances in neural information processing systems},
  author  = {Yeh, Chih-Kuan and Kim, Been and Arik, Sercan and Li, Chun-Liang and Pfister, Tomas and Ravikumar, Pradeep},
  year    = {2020},
  pages   = {20554–20565}
}

@article{Crabbé_van2022,
  title        = {Concept Activation Regions: A Generalized Framework For Concept-Based Explanations},
  url          = {http://arxiv.org/abs/2209.11222},
  doi          = {10.48550/arXiv.2209.11222},
  abstractnote = {Concept-based explanations permit to understand the predictions of a deep neural network (DNN) through the lens of concepts specified by users. Existing methods assume that the examples illustrating a concept are mapped in a fixed direction of the DNN’s latent space. When this holds true, the concept can be represented by a concept activation vector (CAV) pointing in that direction. In this work, we propose to relax this assumption by allowing concept examples to be scattered across different clusters in the DNN’s latent space. Each concept is then represented by a region of the DNN’s latent space that includes these clusters and that we call concept activation region (CAR). To formalize this idea, we introduce an extension of the CAV formalism that is based on the kernel trick and support vector classifiers. This CAR formalism yields global concept-based explanations and local concept-based feature importance. We prove that CAR explanations built with radial kernels are invariant under latent space isometries. In this way, CAR assigns the same explanations to latent spaces that have the same geometry. We further demonstrate empirically that CARs offer (1) more accurate descriptions of how concepts are scattered in the DNN’s latent space; (2) global explanations that are closer to human concept annotations and (3) concept-based feature importance that meaningfully relate concepts with each other. Finally, we use CARs to show that DNNs can autonomously rediscover known scientific concepts, such as the prostate cancer grading system.},
  note         = {arXiv:2209.11222 [cs]},
  number       = {arXiv:2209.11222},
  publisher    = {arXiv},
  author       = {Crabbé, Jonathan and van der Schaar, Mihaela},
  year         = {2022},
  month        = sep
}

@article{Cunningham_Ewart_Riggs_Huben_Sharkey_2023,
  title        = {Sparse Autoencoders Find Highly Interpretable Features in Language Models},
  url          = {http://arxiv.org/abs/2309.08600},
  doi          = {10.48550/arXiv.2309.08600},
  abstractnote = {One of the roadblocks to a better understanding of neural networks’ internals is textit{polysemanticity}, where neurons appear to activate in multiple, semantically distinct contexts. Polysemanticity prevents us from identifying concise, human-understandable explanations for what neural networks are doing internally. One hypothesised cause of polysemanticity is textit{superposition}, where neural networks represent more features than they have neurons by assigning features to an overcomplete set of directions in activation space, rather than to individual neurons. Here, we attempt to identify those directions, using sparse autoencoders to reconstruct the internal activations of a language model. These autoencoders learn sets of sparsely activating features that are more interpretable and monosemantic than directions identified by alternative approaches, where interpretability is measured by automated methods. Moreover, we show that with our learned set of features, we can pinpoint the features that are causally responsible for counterfactual behaviour on the indirect object identification task citep{wang2022interpretability} to a finer degree than previous decompositions. This work indicates that it is possible to resolve superposition in language models using a scalable, unsupervised method. Our method may serve as a foundation for future mechanistic interpretability work, which we hope will enable greater model transparency and steerability.},
  note         = {arXiv:2309.08600 [cs]},
  number       = {arXiv:2309.08600},
  publisher    = {arXiv},
  author       = {Cunningham, Hoagy and Ewart, Aidan and Riggs, Logan and Huben, Robert and Sharkey, Lee},
  year         = {2023},
  month        = oct
}

@misc{darcet2023vitneedreg,
  title   = {Vision Transformers Need Registers},
  author  = {Darcet, Timothée and Oquab, Maxime and Mairal, Julien and Bojanowski, Piotr},
  journal = {arXiv:2309.16588},
  year    = {2023}
}

@inproceedings{DecompX,
  title     = {{D}ecomp{X}: Explaining Transformers Decisions by Propagating Token Decomposition},
  author    = {Modarressi, Ali  and
               Fayyaz, Mohsen  and
               Aghazadeh, Ehsan  and
               Yaghoobzadeh, Yadollah  and
               Pilehvar, Mohammad Taher},
  booktitle = {Proceedings of the 61st Annual Meeting of the Association for Computational Linguistics (Volume 1: Long Papers)},
  month     = jul,
  year      = {2023},
  address   = {Toronto, Canada},
  publisher = {Association for Computational Linguistics},
  url       = {https://aclanthology.org/2023.acl-long.149},
  pages     = {2649--2664}
}

@inproceedings{deng2009imagenet,
  title        = {Imagenet: A large-scale hierarchical image database},
  author       = {Deng, Jia and Dong, Wei and Socher, Richard and Li, Li-Jia and Li, Kai and Fei-Fei, Li},
  booktitle    = {2009 IEEE conference on computer vision and pattern recognition},
  pages        = {248--255},
  year         = {2009},
  organization = {Ieee}
}

@article{devlin2018bert,
  title   = {Bert: Pre-training of deep bidirectional transformers for language understanding},
  author  = {Devlin, Jacob and Chang, Ming-Wei and Lee, Kenton and Toutanova, Kristina},
  journal = {arXiv preprint arXiv:1810.04805},
  year    = {2018},
  note    = {(Apache 2.0)}
}

@article{EAC,
  title        = {Explain Any Concept: Segment Anything Meets Concept-Based Explanation},
  url          = {http://arxiv.org/abs/2305.10289},
  doi          = {10.48550/arXiv.2305.10289},
  abstractnote = {EXplainable AI (XAI) is an essential topic to improve human understanding of deep neural networks (DNNs) given their black-box internals. For computer vision tasks, mainstream pixel-based XAI methods explain DNN decisions by identifying important pixels, and emerging concept-based XAI explore forming explanations with concepts (e.g., a head in an image). However, pixels are generally hard to interpret and sensitive to the imprecision of XAI methods, whereas “concepts” in prior works require human annotation or are limited to pre-defined concept sets. On the other hand, driven by large-scale pre-training, Segment Anything Model (SAM) has been demonstrated as a powerful and promotable framework for performing precise and comprehensive instance segmentation, enabling automatic preparation of concept sets from a given image. This paper for the first time explores using SAM to augment concept-based XAI. We offer an effective and flexible concept-based explanation method, namely Explain Any Concept (EAC), which explains DNN decisions with any concept. While SAM is highly effective and offers an “out-of-the-box” instance segmentation, it is costly when being integrated into defacto XAI pipelines. We thus propose a lightweight per-input equivalent (PIE) scheme, enabling efficient explanation with a surrogate model. Our evaluation over two popular datasets (ImageNet and COCO) illustrate the highly encouraging performance of EAC over commonly-used XAI methods.},
  note         = {arXiv:2305.10289 [cs]},
  number       = {arXiv:2305.10289},
  publisher    = {arXiv},
  author       = {Sun, Ao and Ma, Pingchuan and Yuan, Yuanyuan and Wang, Shuai},
  year         = {2023},
  month        = may
}

@article{Fel_Boutin_Béthune_Cadene_Moayeri_Andéol_Chalvidal_Serre_2023,
  title    = {A Holistic Approach to Unifying Automatic Concept Extraction and Concept Importance Estimation},
  volume   = {36},
  journal  = {Advances in Neural Information Processing Systems},
  author   = {Fel, Thomas and Boutin, Victor and Béthune, Louis and Cadene, Remi and Moayeri, Mazda and Andéol, Léo and Chalvidal, Mathieu and Serre, Thomas},
  year     = {2023},
  month    = dec,
  pages    = {54805–54818},
  language = {en}
}

@misc{geminiteam2024geminifamilyhighlycapable,
  title         = {Gemini: A Family of Highly Capable Multimodal Models},
  author        = {Gemini Team et al.},
  year          = {2024},
  eprint        = {2312.11805},
  archiveprefix = {arXiv},
  primaryclass  = {cs.CL},
  url           = {https://arxiv.org/abs/2312.11805}
}

@article{Ghorbani_Wexler_Zou_Kim_2019,
  title        = {Towards Automatic Concept-based Explanations},
  url          = {http://arxiv.org/abs/1902.03129},
  doi          = {10.48550/arXiv.1902.03129},
  abstractnote = {Interpretability has become an important topic of research as more machine learning (ML) models are deployed and widely used to make important decisions. Most of the current explanation methods provide explanations through feature importance scores, which identify features that are important for each individual input. However, how to systematically summarize and interpret such per sample feature importance scores itself is challenging. In this work, we propose principles and desiderata for emph{concept} based explanation, which goes beyond per-sample features to identify higher-level human-understandable concepts that apply across the entire dataset. We develop a new algorithm, ACE, to automatically extract visual concepts. Our systematic experiments demonstrate that alg discovers concepts that are human-meaningful, coherent and important for the neural network’s predictions.},
  note         = {arXiv:1902.03129 [cs, stat]},
  number       = {arXiv:1902.03129},
  publisher    = {arXiv},
  author       = {Ghorbani, Amirata and Wexler, James and Zou, James and Kim, Been},
  year         = {2019},
  month        = oct
}

@inproceedings{ghorbani2019towards,
  title     = {Towards automatic concept-based explanations},
  author    = {Ghorbani, Amirata and Wexler, James and Zou, James Y and Kim, Been},
  booktitle = {Advances in Neural Information Processing Systems},
  pages     = {9273--9282},
  year      = {2019}
}

@article{GPT4,
  title   = {Gpt-4 technical report},
  author  = {Achiam, Josh and Adler, Steven and Agarwal, Sandhini and Ahmad, Lama and Akkaya, Ilge and Aleman, Florencia Leoni and Almeida, Diogo and Altenschmidt, Janko and Altman, Sam and Anadkat, Shyamal and others},
  journal = {arXiv preprint arXiv:2303.08774},
  year    = {2023}
}

@inproceedings{he2016deep,
  title     = {Deep residual learning for image recognition},
  author    = {He, Kaiming and Zhang, Xiangyu and Ren, Shaoqing and Sun, Jian},
  booktitle = {Proceedings of the IEEE conference on computer vision and pattern recognition},
  pages     = {770--778},
  year      = {2016}
}

@article{ICE,
  author    = {Alex Goldstein and Adam Kapelner and Justin Bleich and Emil Pitkin},
  title     = {Peeking Inside the Black Box: Visualizing Statistical Learning With Plots of Individual Conditional Expectation},
  journal   = {Journal of Computational and Graphical Statistics},
  volume    = {24},
  number    = {1},
  pages     = {44-65},
  year      = {2015},
  publisher = {Taylor & Francis},
  doi       = {10.1080/10618600.2014.907095}
}

@inproceedings{IMDB,
  author    = {Maas, Andrew L.  and  Daly, Raymond E.  and  Pham, Peter T.  and  Huang, Dan  and  Ng, Andrew Y.  and  Potts, Christopher},
  title     = {Learning Word Vectors for Sentiment Analysis},
  booktitle = {Proceedings of the 49th Annual Meeting of the Association for Computational Linguistics: Human Language Technologies},
  month     = {June},
  year      = {2011},
  address   = {Portland, Oregon, USA},
  publisher = {Association for Computational Linguistics},
  pages     = {142--150},
  url       = {http://www.aclweb.org/anthology/P11-1015}
}

@article{ismail2021improving,
  title   = {Improving deep learning interpretability by saliency guided training},
  author  = {Ismail, Aya Abdelsalam and Corrada Bravo, Hector and Feizi, Soheil},
  journal = {Advances in Neural Information Processing Systems},
  volume  = {34},
  pages   = {26726--26739},
  year    = {2021}
}

@misc{jiang2018quickshiftprovablygoodinitializations,
  title         = {Quickshift++: Provably Good Initializations for Sample-Based Mean Shift},
  author        = {Heinrich Jiang and Jennifer Jang and Samory Kpotufe},
  year          = {2018},
  eprint        = {1805.07909},
  archiveprefix = {arXiv},
  primaryclass  = {cs.LG},
  url           = {https://arxiv.org/abs/1805.07909}
}

@inproceedings{Kim_Wattenberg_Gilmer_Cai_Wexler_Viegas_2018,
  title     = {Interpretability beyond feature attribution: Quantitative testing with concept activation vectors (tcav)},
  url       = {http://proceedings.mlr.press/v80/kim18d.html},
  booktitle = {International conference on machine learning},
  publisher = {PMLR},
  author    = {Kim, Been and Wattenberg, Martin and Gilmer, Justin and Cai, Carrie and Wexler, James and Viegas, Fernanda},
  year      = {2018},
  pages     = {2668–2677}
}

@article{kirillov2023segany,
  title   = {Segment Anything},
  author  = {Kirillov, Alexander and Mintun, Eric and Ravi, Nikhila and Mao, Hanzi and Rolland, Chloe and Gustafson, Laura and Xiao, Tete and Whitehead, Spencer and Berg, Alexander C. and Lo, Wan-Yen and Doll{\'a}r, Piotr and Girshick, Ross},
  journal = {arXiv:2304.02643},
  year    = {2023}
}

@inproceedings{KL-LUCB,
  title        = {Information complexity in bandit subset selection},
  author       = {Kaufmann, Emilie and Kalyanakrishnan, Shivaram},
  booktitle    = {Conference on Learning Theory},
  pages        = {228--251},
  year         = {2013},
  organization = {PMLR}
}

@inproceedings{LIME,
  author    = {Marco T{\'{u}}lio Ribeiro and
               Sameer Singh and
               Carlos Guestrin},
  editor    = {Balaji Krishnapuram and
               Mohak Shah and
               Alexander J. Smola and
               Charu C. Aggarwal and
               Dou Shen and
               Rajeev Rastogi},
  title     = {``Why Should {I} Trust You?": Explaining the Predictions of Any Classifier},
  booktitle = {Proceedings of the 22nd {ACM} {SIGKDD} International Conference on
               Knowledge Discovery and Data Mining, San Francisco, CA, USA, August
               13-17, 2016},
  pages     = {1135--1144},
  publisher = {{ACM}},
  year      = {2016}
}

@article{lore,
  author     = {Riccardo Guidotti and
                Anna Monreale and
                Salvatore Ruggieri and
                Dino Pedreschi and
                Franco Turini and
                Fosca Giannotti},
  title      = {Local Rule-Based Explanations of Black Box Decision Systems},
  journal    = {CoRR},
  volume     = {abs/1805.10820},
  year       = {2018},
  url        = {http://arxiv.org/abs/1805.10820},
  eprinttype = {arXiv},
  eprint     = {1805.10820},
  bibsource  = {dblp computer science bibliography, https://dblp.org}
}

@book{molnar2020interpretable,
  title     = {Interpretable machine learning},
  author    = {Molnar, Christoph},
  year      = {2020},
  publisher = {Lulu. com}
}

@misc{oquab2023dinov2,
  title   = {DINOv2: Learning Robust Visual Features without Supervision},
  author  = {Oquab, Maxime and Darcet, Timothée and Moutakanni, Theo and Vo, Huy V. and Szafraniec, Marc and Khalidov, Vasil and Fernandez, Pierre and Haziza, Daniel and Massa, Francisco and El-Nouby, Alaaeldin and Howes, Russell and Huang, Po-Yao and Xu, Hu and Sharma, Vasu and Li, Shang-Wen and Galuba, Wojciech and Rabbat, Mike and Assran, Mido and Ballas, Nicolas and Synnaeve, Gabriel and Misra, Ishan and Jegou, Herve and Mairal, Julien and Labatut, Patrick and Joulin, Armand and Bojanowski, Piotr},
  journal = {arXiv:2304.07193},
  year    = {2023}
}

@inproceedings{ReX,
  title={ReX: A framework for incorporating temporal information in model-agnostic local explanation techniques},
  author={Liu, Junhao and Zhang, Xin},
  booktitle={Proceedings of the AAAI Conference on Artificial Intelligence},
  volume={39},
  number={18},
  pages={18888--18896},
  year={2025}
}

@article{samek2016evaluating,
  title     = {Evaluating the visualization of what a deep neural network has learned},
  author    = {Samek, Wojciech and Binder, Alexander and Montavon, Gr{\'e}goire and Lapuschkin, Sebastian and M{\"u}ller, Klaus-Robert},
  journal   = {IEEE transactions on neural networks and learning systems},
  volume    = {28},
  number    = {11},
  pages     = {2660--2673},
  year      = {2016},
  publisher = {IEEE}
}

@article{Shankaranarayana_Runje_2019,
  title        = {ALIME: Autoencoder Based Approach for Local Interpretability},
  url          = {http://arxiv.org/abs/1909.02437},
  doi          = {10.48550/arXiv.1909.02437},
  abstractnote = {Machine learning and especially deep learning have garneredtremendous popularity in recent years due to their increased performanceover other methods. The availability of large amount of data has aidedin the progress of deep learning. Nevertheless, deep learning models areopaque and often seen as black boxes. Thus, there is an inherent need tomake the models interpretable, especially so in the medical domain. Inthis work, we propose a locally interpretable method, which is inspiredby one of the recent tools that has gained a lot of interest, called localinterpretable model-agnostic explanations (LIME). LIME generates singleinstance level explanation by artificially generating a dataset aroundthe instance (by randomly sampling and using perturbations) and thentraining a local linear interpretable model. One of the major issues inLIME is the instability in the generated explanation, which is caused dueto the randomly generated dataset. Another issue in these kind of localinterpretable models is the local fidelity. We propose novel modificationsto LIME by employing an autoencoder, which serves as a better weightingfunction for the local model. We perform extensive comparisons withdifferent datasets and show that our proposed method results in bothimproved stability, as well as local fidelity.},
  note         = {arXiv:1909.02437 [cs, stat]},
  number       = {arXiv:1909.02437},
  publisher    = {arXiv},
  author       = {Shankaranarayana, Sharath M. and Runje, Davor},
  year         = {2019},
  month        = sep
}

@inproceedings{SHAP,
  author    = {Scott M. Lundberg and
               Su{-}In Lee},
  editor    = {Isabelle Guyon and
               Ulrike von Luxburg and
               Samy Bengio and
               Hanna M. Wallach and
               Rob Fergus and
               S. V. N. Vishwanathan and
               Roman Garnett},
  title     = {A Unified Approach to Interpreting Model Predictions},
  booktitle = {Advances in Neural Information Processing Systems 30: Annual Conference
               on Neural Information Processing Systems 2017, December 4-9, 2017,
               Long Beach, CA, {USA}},
  pages     = {4765--4774},
  year      = {2017}
}

@article{survey_model_agnostic,
  title        = {A Comparative Analysis of Model Agnostic Techniques for Explainable Artificial Intelligence},
  rights       = {Copyright (c) 2024 Yifei Wang},
  issn         = {2811-0013},
  doi          = {10.37256/rrcs.3220244750},
  abstractnote = {Explainable Artificial Intelligence (XAI) has become essential as AI systems increasingly influence critical domains, demanding transparency for trust and validation. This paper presents a comparative analysis of prominent model agnostic techniques designed to provide interpretability irrespective of the underlying model architecture. We explore Local Interpretable Model-agnostic Explanations (LIME), SHapley Additive exPlanations (SHAP), Partial Dependence Plots (PDP), Individual Conditional Expectation (ICE) plots, and Anchors. Our analysis focuses on several criteria including interpretative clarity, computational efficiency, scalability, and user-friendliness. Results indicate significant differences in the applicability of each technique depending on the complexity and type of data, highlighting SHAP and LIME for their robustness and detailed output, whereas PDP and ICE are noted for their simplicity in usage and interpretation. The study emphasizes the importance of context in choosing appropriate XAI techniques and suggests directions for future research to enhance the efficacy of model agnostic approaches in explainability. This work contributes to a deeper understanding of how different XAI techniques can be effectively deployed in practice, guiding developers and researchers in making informed decisions about implementing AI transparency.},
  journal      = {Research Reports on Computer Science},
  author       = {Wang, Yifei},
  year         = {2024},
  month        = aug,
  pages        = {25–33},
  language     = {en}
}

@article{survey_zhang,
  title        = {A Survey on Neural Network Interpretability},
  volume       = {5},
  issn         = {2471-285X},
  doi          = {10.1109/TETCI.2021.3100641},
  abstractnote = {Along with the great success of deep neural networks, there is also growing concern about their black-box nature. The interpretability issue affects people’s trust on deep learning systems. It is also related to many ethical problems, e.g., algorithmic discrimination. Moreover, interpretability is a desired property for deep networks to become powerful tools in other research fields, e.g., drug discovery and genomics. In this survey, we conduct a comprehensive review of the neural network interpretability research. We first clarify the definition of interpretability as it has been used in many different contexts. Then we elaborate on the importance of interpretability and propose a novel taxonomy organized along three dimensions: type of engagement (passive vs. active interpretation approaches), the type of explanation, and the focus (from local to global interpretability). This taxonomy provides a meaningful 3D view of distribution of papers from the relevant literature as two of the dimensions are not simply categorical but allow ordinal subcategories. Finally, we summarize the existing interpretability evaluation methods and suggest possible research directions inspired by our new taxonomy.},
  note         = {arXiv:2012.14261 [cs]},
  number       = {5},
  journal      = {IEEE Transactions on Emerging Topics in Computational Intelligence},
  author       = {Zhang, Yu and Tiňo, Peter and Leonardis, Aleš and Tang, Ke},
  year         = {2021},
  month        = oct,
  pages        = {726–742}
}

@misc{tan2023glime,
  title         = {GLIME: General, Stable and Local LIME Explanation},
  author        = {Zeren Tan and Yang Tian and Jian Li},
  year          = {2023},
  eprint        = {2311.15722},
  archiveprefix = {arXiv},
  primaryclass  = {cs.LG}
}

@article{TBM,
  title        = {Interpretable-by-Design Text Classification with Iteratively Generated Concept Bottleneck},
  url          = {http://arxiv.org/abs/2310.19660},
  doi          = {10.48550/arXiv.2310.19660},
  abstractnote = {Deep neural networks excel in text classification tasks, yet their application in high-stakes domains is hindered by their lack of interpretability. To address this, we propose Text Bottleneck Models (TBMs), an intrinsically interpretable text classification framework that offers both global and local explanations. Rather than directly predicting the output label, TBMs predict categorical values for a sparse set of salient concepts and use a linear layer over those concept values to produce the final prediction. These concepts can be automatically discovered and measured by a Large Language Model (LLM), without the need for human curation. On 12 diverse datasets, using GPT-4 for both concept generation and measurement, we show that TBMs can rival the performance of established black-box baselines such as GPT-4 fewshot and finetuned DeBERTa, while falling short against finetuned GPT-3.5. Overall, our findings suggest that TBMs are a promising new framework that enhances interpretability, with minimal performance tradeoffs, particularly for general-domain text.},
  note         = {arXiv:2310.19660 [cs]},
  number       = {arXiv:2310.19660},
  publisher    = {arXiv},
  author       = {Ludan, Josh Magnus and Lyu, Qing and Yang, Yue and Dugan, Liam and Yatskar, Mark and Callison-Burch, Chris},
  year         = {2023},
  month        = oct
}

@misc{wachter2018counterfactualexplanationsopeningblack,
  title         = {Counterfactual Explanations without Opening the Black Box: Automated Decisions and the GDPR},
  author        = {Sandra Wachter and Brent Mittelstadt and Chris Russell},
  year          = {2018},
  eprint        = {1711.00399},
  archiveprefix = {arXiv},
  primaryclass  = {cs.AI},
  url           = {https://arxiv.org/abs/1711.00399}
}

@misc{widmer2022humancompatiblexaiexplainingdata,
  title         = {Towards Human-Compatible XAI: Explaining Data Differentials with Concept Induction over Background Knowledge},
  author        = {Cara Widmer and Md Kamruzzaman Sarker and Srikanth Nadella and Joshua Fiechter and Ion Juvina and Brandon Minnery and Pascal Hitzler and Joshua Schwartz and Michael Raymer},
  year          = {2022},
  eprint        = {2209.13710},
  archiveprefix = {arXiv},
  primaryclass  = {cs.AI},
  url           = {https://arxiv.org/abs/2209.13710}
}

@inproceedings{YADT,
  title     = {Yadt: Yet another decision tree builder},
  url       = {https://ieeexplore.ieee.org/abstract/document/1374196/},
  booktitle = {16th IEEE International Conference on Tools with Artificial Intelligence},
  publisher = {IEEE},
  author    = {Ruggieri, Salvatore},
  year      = {2004},
  pages     = {260–265}
}

@article{Yeh_Kim_Arik_Li_Ravikumar_Pfister_2019,
  title        = {On Concept-Based Explanations in Deep Neural Networks},
  url          = {https://openreview.net/forum?id=BylWYC4KwH},
  abstractnote = {Deep neural networks (DNNs) build high-level intelligence on low-level raw features. Understanding of this high-level intelligence can be enabled by deciphering the concepts they base their decisions on, as human-level thinking. In this paper, we study concept-based explainability for DNNs in a systematic framework. First, we define the notion of completeness, which quantifies how sufficient a particular set of concepts is in explaining a model’s prediction behavior. Based on performance and variability motivations, we propose two definitions to quantify completeness. We show that under degenerate conditions, our method is equivalent to Principal Component Analysis. Next, we propose a concept discovery method that considers two additional constraints to encourage the interpretability of the discovered concepts. We use game-theoretic notions to aggregate over sets to define an importance score for each discovered concept, which we call emph{ConceptSHAP}. On specifically-designed synthetic datasets and real-world text and image datasets, we validate the effectiveness of our framework in finding concepts that are complete in explaining the decision, and interpretable.},
  author       = {Yeh, Chih-Kuan and Kim, Been and Arik, Sercan and Li, Chun-Liang and Ravikumar, Pradeep and Pfister, Tomas},
  year         = {2019},
  month        = sep,
  language     = {en}
}

@article{Zhang_Madumal_Miller_Ehinger_Rubinstein_2021,
  title        = {Invertible Concept-based Explanations for CNN Models with Non-negative Concept Activation Vectors},
  url          = {http://arxiv.org/abs/2006.15417},
  doi          = {10.48550/arXiv.2006.15417},
  abstractnote = {Convolutional neural network (CNN) models for computer vision are powerful but lack explainability in their most basic form. This deficiency remains a key challenge when applying CNNs in important domains. Recent work on explanations through feature importance of approximate linear models has moved from input-level features (pixels or segments) to features from mid-layer feature maps in the form of concept activation vectors (CAVs). CAVs contain concept-level information and could be learned via clustering. In this work, we rethink the ACE algorithm of Ghorbani et~al., proposing an alternative invertible concept-based explanation (ICE) framework to overcome its shortcomings. Based on the requirements of fidelity (approximate models to target models) and interpretability (being meaningful to people), we design measurements and evaluate a range of matrix factorization methods with our framework. We find that non-negative concept activation vectors (NCAVs) from non-negative matrix factorization provide superior performance in interpretability and fidelity based on computational and human subject experiments. Our framework provides both local and global concept-level explanations for pre-trained CNN models.},
  note         = {arXiv:2006.15417 [cs]},
  number       = {arXiv:2006.15417},
  publisher    = {arXiv},
  author       = {Zhang, Ruihan and Madumal, Prashan and Miller, Tim and Ehinger, Krista A. and Rubinstein, Benjamin I. P.},
  year         = {2021},
  month        = jun
}

@misc{zhang2021invertibleconceptbasedexplanationscnn,
  title         = {Invertible Concept-based Explanations for CNN Models with Non-negative Concept Activation Vectors},
  author        = {Ruihan Zhang and Prashan Madumal and Tim Miller and Krista A. Ehinger and Benjamin I. P. Rubinstein},
  year          = {2021},
  eprint        = {2006.15417},
  archiveprefix = {arXiv},
  primaryclass  = {cs.CV},
  url           = {https://arxiv.org/abs/2006.15417}
}

@article{LDM,
author = {Avrahami, Omri and Fried, Ohad and Lischinski, Dani},
title = {Blended Latent Diffusion},
year = {2023},
issue_date = {August 2023},
publisher = {Association for Computing Machinery},
address = {New York, NY, USA},
volume = {42},
number = {4},
issn = {0730-0301},
url = {https://doi.org/10.1145/3592450},
doi = {10.1145/3592450},
journal = {ACM Trans. Graph.},
month = jul,
articleno = {149},
numpages = {11}
}

@inproceedings{fake-news-dataset,
    title = "Automatic Detection of Fake News",
    author = "P{\'e}rez-Rosas, Ver{\'o}nica  and
      Kleinberg, Bennett  and
      Lefevre, Alexandra  and
      Mihalcea, Rada",
    editor = "Bender, Emily M.  and
      Derczynski, Leon  and
      Isabelle, Pierre",
    booktitle = "Proceedings of the 27th International Conference on Computational Linguistics",
    month = aug,
    year = "2018",
    address = "Santa Fe, New Mexico, USA",
    publisher = "Association for Computational Linguistics",
    url = "https://aclanthology.org/C18-1287/",
    pages = "3391--3401"
}

@article{COCO,
  author       = {Tsung{-}Yi Lin and
                  Michael Maire and
                  Serge J. Belongie and
                  Lubomir D. Bourdev and
                  Ross B. Girshick and
                  James Hays and
                  Pietro Perona and
                  Deva Ramanan and
                  Piotr Doll{\'{a}}r and
                  C. Lawrence Zitnick},
  title        = {Microsoft {COCO:} Common Objects in Context},
  journal      = {CoRR},
  volume       = {abs/1405.0312},
  year         = {2014},
  url          = {http://arxiv.org/abs/1405.0312},
  eprinttype    = {arXiv},
  eprint       = {1405.0312},
  bibsource    = {dblp computer science bibliography, https://dblp.org}
}

@software{YOLO11,
author = {Jocher, Glenn and Qiu, Jing and Chaurasia, Ayush},
license = {AGPL-3.0},
month = jan,
title = {{Ultralytics YOLO}},
url = {https://github.com/ultralytics/ultralytics},
version = {8.0.0},
year = {2023}
}

@article{liu2024deepseekv3,
  title={Deepseek-v3 technical report},
  author={Liu, Aixin and Feng, Bei and Xue, Bing and Wang, Bingxuan and Wu, Bochao and Lu, Chengda and Zhao, Chenggang and Deng, Chengqi and Zhang, Chenyu and Ruan, Chong and others},
  journal={arXiv preprint arXiv:2412.19437},
  year={2024}
}

@article{Fel_Picard_Bethune_Boissin_Vigouroux_Colin_Cadène_Serre_2023, title={CRAFT: Concept Recursive Activation FacTorization for Explainability}, url={http://arxiv.org/abs/2211.10154}, DOI={10.48550/arXiv.2211.10154}, abstractNote={Attribution methods, which employ heatmaps to identify the most influential regions of an image that impact model decisions, have gained widespread popularity as a type of explainability method. However, recent research has exposed the limited practical value of these methods, attributed in part to their narrow focus on the most prominent regions of an image -- revealing “where” the model looks, but failing to elucidate “what” the model sees in those areas. In this work, we try to fill in this gap with CRAFT -- a novel approach to identify both “what” and “where” by generating concept-based explanations. We introduce 3 new ingredients to the automatic concept extraction literature: (i) a recursive strategy to detect and decompose concepts across layers, (ii) a novel method for a more faithful estimation of concept importance using Sobol indices, and (iii) the use of implicit differentiation to unlock Concept Attribution Maps. We conduct both human and computer vision experiments to demonstrate the benefits of the proposed approach. We show that the proposed concept importance estimation technique is more faithful to the model than previous methods. When evaluating the usefulness of the method for human experimenters on a human-centered utility benchmark, we find that our approach significantly improves on two of the three test scenarios. Our code is freely available at github.com/deel-ai/Craft.}, note={arXiv:2211.10154}, number={arXiv:2211.10154}, publisher={arXiv}, author={Fel, Thomas and Picard, Agustin and Bethune, Louis and Boissin, Thibaut and Vigouroux, David and Colin, Julien and Cadène, Rémi and Serre, Thomas}, year={2023}, month=mar }

@article{caltech101,
  title={Learning generative visual models from few training examples: An incremental Bayesian approach tested on 101 object categories},
  author={Fei-Fei, Li and Fergus, Rob and Perona, Pietro},
  journal={Computer vision and Image understanding},
  volume={106},
  number={1},
  pages={59--70},
  year={2007},
  publisher={Elsevier}
}

@article{ACE, title={Towards Automatic Concept-based Explanations}, url={http://arxiv.org/abs/1902.03129}, DOI={10.48550/arXiv.1902.03129}, abstractNote={Interpretability has become an important topic of research as more machine learning (ML) models are deployed and widely used to make important decisions. Most of the current explanation methods provide explanations through feature importance scores, which identify features that are important for each individual input. However, how to systematically summarize and interpret such per sample feature importance scores itself is challenging. In this work, we propose principles and desiderata for emph{concept} based explanation, which goes beyond per-sample features to identify higher-level human-understandable concepts that apply across the entire dataset. We develop a new algorithm, ACE, to automatically extract visual concepts. Our systematic experiments demonstrate that alg discovers concepts that are human-meaningful, coherent and important for the neural network’s predictions.}, note={arXiv:1902.03129 [cs, stat]}, number={arXiv:1902.03129}, publisher={arXiv}, author={Ghorbani, Amirata and Wexler, James and Zou, James and Kim, Been}, year={2019}, month=oct }

@article{mcdonald2009ridge,
  title={Ridge regression},
  author={McDonald, Gary C},
  journal={Wiley Interdisciplinary Reviews: Computational Statistics},
  volume={1},
  number={1},
  pages={93--100},
  year={2009},
  publisher={Wiley Online Library}
}

@article{Oikarinen_Das_Nguyen_Weng_2023, title={Label-Free Concept Bottleneck Models}, url={http://arxiv.org/abs/2304.06129}, DOI={10.48550/arXiv.2304.06129}, abstractNote={Concept bottleneck models (CBM) are a popular way of creating more interpretable neural networks by having hidden layer neurons correspond to human-understandable concepts. However, existing CBMs and their variants have two crucial limitations: first, they need to collect labeled data for each of the predefined concepts, which is time consuming and labor intensive; second, the accuracy of a CBM is often significantly lower than that of a standard neural network, especially on more complex datasets. This poor performance creates a barrier for adopting CBMs in practical real world applications. Motivated by these challenges, we propose Label-free CBM which is a novel framework to transform any neural network into an interpretable CBM without labeled concept data, while retaining a high accuracy. Our Label-free CBM has many advantages, it is: scalable - we present the first CBM scaled to ImageNet, efficient - creating a CBM takes only a few hours even for very large datasets, and automated - training it for a new dataset requires minimal human effort. Our code is available at https://github.com/Trustworthy-ML-Lab/Label-free-CBM. Finally, in Appendix B we conduct a large scale user evaluation of the interpretability of our method.}, note={arXiv:2304.06129 [cs]}, number={arXiv:2304.06129}, publisher={arXiv}, author={Oikarinen, Tuomas and Das, Subhro and Nguyen, Lam M. and Weng, Tsui-Wei}, year={2023}, month=jun }

@inproceedings{Koh_Nguyen_Tang_Mussmann_Pierson_Kim_Liang_2020, title={Concept Bottleneck Models}, ISSN={2640-3498}, url={https://proceedings.mlr.press/v119/koh20a.html}, abstractNote={We seek to learn models that we can interact with using high-level concepts: if the model did not think there was a bone spur in the x-ray, would it still predict severe arthritis? State-of-the-art models today do not typically support the manipulation of concepts like “the existence of bone spurs”, as they are trained end-to-end to go directly from raw input (e.g., pixels) to output (e.g., arthritis severity). We revisit the classic idea of first predicting concepts that are provided at training time, and then using these concepts to predict the label. By construction, we can intervene on these concept bottleneck models by editing their predicted concept values and propagating these changes to the final prediction. On x-ray grading and bird identification, concept bottleneck models achieve competitive accuracy with standard end-to-end models, while enabling interpretation in terms of high-level clinical concepts (“bone spurs”) or bird attributes (“wing color”). These models also allow for richer human-model interaction: accuracy improves significantly if we can correct model mistakes on concepts at test time.}, booktitle={Proceedings of the 37th International Conference on Machine Learning}, publisher={PMLR}, author={Koh, Pang Wei and Nguyen, Thao and Tang, Yew Siang and Mussmann, Stephen and Pierson, Emma and Kim, Been and Liang, Percy}, year={2020}, month=nov, pages={5338–5348}, language={en} }

@article{Kim_Jung_Park_Kim_Yoon_2023, title={Probabilistic Concept Bottleneck Models}, url={http://arxiv.org/abs/2306.01574}, DOI={10.48550/arXiv.2306.01574}, abstractNote={Interpretable models are designed to make decisions in a human-interpretable manner. Representatively, Concept Bottleneck Models (CBM) follow a two-step process of concept prediction and class prediction based on the predicted concepts. CBM provides explanations with high-level concepts derived from concept predictions; thus, reliable concept predictions are important for trustworthiness. In this study, we address the ambiguity issue that can harm reliability. While the existence of a concept can often be ambiguous in the data, CBM predicts concepts deterministically without considering this ambiguity. To provide a reliable interpretation against this ambiguity, we propose Probabilistic Concept Bottleneck Models (ProbCBM). By leveraging probabilistic concept embeddings, ProbCBM models uncertainty in concept prediction and provides explanations based on the concept and its corresponding uncertainty. This uncertainty enhances the reliability of the explanations. Furthermore, as class uncertainty is derived from concept uncertainty in ProbCBM, we can explain class uncertainty by means of concept uncertainty. Code is publicly available at https://github.com/ejkim47/prob-cbm.}, note={arXiv:2306.01574}, number={arXiv:2306.01574}, publisher={arXiv}, author={Kim, Eunji and Jung, Dahuin and Park, Sangha and Kim, Siwon and Yoon, Sungroh}, year={2023}, month=jun }

@inproceedings{tjong-kim-sang-de-meulder-2003-introduction,
    title = "Introduction to the {C}o{NLL}-2003 Shared Task: Language-Independent Named Entity Recognition",
    author = "Tjong Kim Sang, Erik F.  and
      De Meulder, Fien",
    booktitle = "Proceedings of the Seventh Conference on Natural Language Learning at {HLT}-{NAACL} 2003",
    year = "2003",
    url = "https://www.aclweb.org/anthology/W03-0419",
    pages = "142--147",
}

@inproceedings{pradhan2013towards,
  title={Towards robust linguistic analysis using OntoNotes},
  author={Pradhan, Sameer S and Hovy, Eduard and Marcus, Mitchell and Palmer, Martha and Ramshaw, Lance and Weischedel, Ralph},
  booktitle={Proceedings of the Seventeenth Conference on Computational Natural Language Learning},
  pages={143--152},
  year={2013}
}

@inproceedings{Goyal2017Making,
  title={Making the V in VQA Matter: Elevating the Role of Image Understanding in Visual Question Answering},
  author={Yash Goyal and Tejas Khot and Douglas Summers-Stay and Dhruv Batra and Devi Parikh},
  booktitle={Proceedings of the IEEE Conference on Computer Vision and Pattern Recognition (CVPR)},
  year={2017}
}

@misc{warren2022featuresexplainabilityusersunderstand,
      title={Features of Explainability: How users understand counterfactual and causal explanations for categorical and continuous features in XAI}, 
      author={Greta Warren and Mark T Keane and Ruth M J Byrne},
      year={2022},
      eprint={2204.10152},
      archivePrefix={arXiv},
      primaryClass={cs.HC},
      url={https://arxiv.org/abs/2204.10152}, 
}

@inproceedings{hase-bansal-2020-evaluating,
    title = "Evaluating Explainable {AI}: Which Algorithmic Explanations Help Users Predict Model Behavior?",
    author = "Hase, Peter  and
      Bansal, Mohit",
    editor = "Jurafsky, Dan  and
      Chai, Joyce  and
      Schluter, Natalie  and
      Tetreault, Joel",
    booktitle = "Proceedings of the 58th Annual Meeting of the Association for Computational Linguistics",
    month = jul,
    year = "2020",
    address = "Online",
    publisher = "Association for Computational Linguistics",
    url = "https://aclanthology.org/2020.acl-main.491/",
    doi = "10.18653/v1/2020.acl-main.491",
    pages = "5540--5552"
}

@article{welinder2010caltech,
  title={Caltech-UCSD birds 200},
  author={Welinder, Peter and Branson, Steve and Mita, Takeshi and Wah, Catherine and Schroff, Florian and Belongie, Serge and Perona, Pietro},
  year={2010},
  publisher={Technical Report CNS-TR-2010-001, California Institute of Technology}
}

@article{Ciravegna_Barbiero_Giannini_Gori_Lió_Maggini_Melacci_2023, title={Logic Explained Networks}, volume={314}, ISSN={0004-3702}, DOI={10.1016/j.artint.2022.103822}, abstractNote={The large and still increasing popularity of deep learning clashes with a major limit of neural network architectures, that consists in their lack of capability in providing human-understandable motivations of their decisions. In situations in which the machine is expected to support the decision of human experts, providing a comprehensible explanation is a feature of crucial importance. The language used to communicate the explanations must be formal enough to be implementable in a machine and friendly enough to be understandable by a wide audience. In this paper, we propose a general approach to Explainable Artificial Intelligence in the case of neural architectures, showing how a mindful design of the networks leads to a family of interpretable deep learning models called Logic Explained Networks (LENs). LENs only require their inputs to be human-understandable predicates, and they provide explanations in terms of simple First-Order Logic (FOL) formulas involving such predicates. LENs are general enough to cover a large number of scenarios. Amongst them, we consider the case in which LENs are directly used as special classifiers with the capability of being explainable, or when they act as additional networks with the role of creating the conditions for making a black-box classifier explainable by FOL formulas. Despite supervised learning problems are mostly emphasized, we also show that LENs can learn and provide explanations in unsupervised learning settings. Experimental results on several datasets and tasks show that LENs may yield better classifications than established white-box models, such as decision trees and Bayesian rule lists, while providing more compact and meaningful explanations.}, note={arXiv:2108.05149 [cs]}, journal={Artificial Intelligence}, author={Ciravegna, Gabriele and Barbiero, Pietro and Giannini, Francesco and Gori, Marco and Lió, Pietro and Maggini, Marco and Melacci, Stefano}, year={2023}, month=jan, pages={103822} }

@article{Goyal_Feder_Shalit_Kim_2020, title={Explaining Classifiers with Causal Concept Effect (CaCE)}, url={http://arxiv.org/abs/1907.07165}, DOI={10.48550/arXiv.1907.07165}, abstractNote={How can we understand classification decisions made by deep neural networks? Many existing explainability methods rely solely on correlations and fail to account for confounding, which may result in potentially misleading explanations. To overcome this problem, we define the Causal Concept Effect (CaCE) as the causal effect of (the presence or absence of) a human-interpretable concept on a deep neural net’s predictions. We show that the CaCE measure can avoid errors stemming from confounding. Estimating CaCE is difficult in situations where we cannot easily simulate the do-operator. To mitigate this problem, we use a generative model, specifically a Variational AutoEncoder (VAE), to measure VAE-CaCE. In an extensive experimental analysis, we show that the VAE-CaCE is able to estimate the true concept causal effect, compared to baselines for a number of datasets including high dimensional images.}, note={arXiv:1907.07165 [cs]}, number={arXiv:1907.07165}, publisher={arXiv}, author={Goyal, Yash and Feder, Amir and Shalit, Uri and Kim, Been}, year={2020}, month=feb }

@article{Wu_D’Oosterlinck_Geiger_Zur_Potts_2022, title={Causal Proxy Models for Concept-Based Model Explanations}, url={http://arxiv.org/abs/2209.14279}, DOI={10.48550/arXiv.2209.14279}, abstractNote={Explainability methods for NLP systems encounter a version of the fundamental problem of causal inference: for a given ground-truth input text, we never truly observe the counterfactual texts necessary for isolating the causal effects of model representations on outputs. In response, many explainability methods make no use of counterfactual texts, assuming they will be unavailable. In this paper, we show that robust causal explainability methods can be created using approximate counterfactuals, which can be written by humans to approximate a specific counterfactual or simply sampled using metadata-guided heuristics. The core of our proposal is the Causal Proxy Model (CPM). A CPM explains a black-box model $mathcal{N}$ because it is trained to have the same actual input/output behavior as $mathcal{N}$ while creating neural representations that can be intervened upon to simulate the counterfactual input/output behavior of $mathcal{N}$. Furthermore, we show that the best CPM for $mathcal{N}$ performs comparably to $mathcal{N}$ in making factual predictions, which means that the CPM can simply replace $mathcal{N}$, leading to more explainable deployed models. Our code is available at https://github.com/frankaging/Causal-Proxy-Model.}, note={arXiv:2209.14279 [cs]}, number={arXiv:2209.14279}, publisher={arXiv}, author={Wu, Zhengxuan and D’Oosterlinck, Karel and Geiger, Atticus and Zur, Amir and Potts, Christopher}, year={2022}, month=sept }

@article{yu2024latent,
  title={Latent concept-based explanation of nlp models},
  author={Yu, Xuemin and Dalvi, Fahim and Durrani, Nadir and Nouri, Marzia and Sajjad, Hassan},
  journal={arXiv preprint arXiv:2404.12545},
  year={2024}
}

@inproceedings{tan2024concept,
  title={A concept-based local interpretable model-agnostic explanation approach for deep neural networks in image classification},
  author={Tan, Lidan and Huang, Changwu and Yao, Xin},
  booktitle={International Conference on Intelligent Information Processing},
  pages={119--133},
  year={2024},
  organization={Springer}
}

@misc{varshney2025generatingcounterfactualtrajectorieslatent,
      title={Generating Counterfactual Trajectories with Latent Diffusion Models for Concept Discovery}, 
      author={Payal Varshney and Adriano Lucieri and Christoph Balada and Andreas Dengel and Sheraz Ahmed},
      year={2025},
      eprint={2404.10356},
      archivePrefix={arXiv},
      primaryClass={cs.LG},
      url={https://arxiv.org/abs/2404.10356}, 
}

@inproceedings{tapariaexplainable,
  title={Explainable Concept Generation through Vision-Language Preference Learning for Understanding Neural Networks' Internal Representations},
  author={Taparia, Aditya and Sagar, Som and Senanayake, Ransalu},
  booktitle={Forty-second International Conference on Machine Learning}
}

@misc{LCM,
      title={LCM-LoRA: A Universal Stable-Diffusion Acceleration Module}, 
      author={Simian Luo and Yiqin Tan and Suraj Patil and Daniel Gu and Patrick von Platen and Apolinário Passos and Longbo Huang and Jian Li and Hang Zhao},
      year={2023},
      eprint={2311.05556},
      archivePrefix={arXiv},
      primaryClass={cs.CV},
      url={https://arxiv.org/abs/2311.05556}, 
}

@misc{HIPIE,
      title={Hierarchical Open-vocabulary Universal Image Segmentation}, 
      author={Xudong Wang and Shufan Li and Konstantinos Kallidromitis and Yusuke Kato and Kazuki Kozuka and Trevor Darrell},
      year={2023},
      eprint={2307.00764},
      archivePrefix={arXiv},
      primaryClass={cs.CV},
      url={https://arxiv.org/abs/2307.00764}, 
}

@inproceedings{bravo2023open,
  title={Open-vocabulary attribute detection},
  author={Bravo, Maria A and Mittal, Sudhanshu and Ging, Simon and Brox, Thomas},
  booktitle={Proceedings of the IEEE/CVF conference on computer vision and pattern recognition},
  pages={7041--7050},
  year={2023}
}
\bibliographystyle{icml2026}

\newpage
\appendix
\onecolumn
\clearpage
\appendix

\section{The Use of Large Language Models}
We use LLMs to refine and polish human writing, and find related work with DeepResearch. We do not use LLMs to generate the main content or ideas of this paper.

\section{The \toolname Framework (Continued)}

\label{app:framework}
In this section, we introduce the details of incorporating \toolname into existing local explanation methods. 
%

We first follow Section~\ref{sec:fram} to introduce how we extend each part for text models in detail.
\subsection{Producing Concept}

\toolname provides predicates that describe high-level concepts \textbf{following existing concept-based methods}.

Specifically, we follow~\citet{TBM} for text data and~\citet{EAC} for image data, and keep all theirs settings.

\subsection{Concept-Feature Mapping}

\textbf{Text Data.}
We the the following prompt for predicate-feature mapping:
\begin{quote}
Generate a sentence similar to a given sentence from the domain of \{\} dataset. The dataset's description is that \{\}.\\

The generated sentence satisfies given concepts. Before generating the sentence, carefully read the description of each concept to understand the properties that the generated sentence must satisfy, think about how the sentence satisfies the concepts first, and then create the sentence that satisfies the concepts.\\

Format your response as a JSON with string keys and string values. Below is an example of a valid JSON response. The JSON contains keys \texttt{thoughts}, and \texttt{answer}. End your response with \textbackslash\#\#\#\\

\noindent\rule{\linewidth}{0.4pt}\\
Concepts:\\
1. Concept 1\\
2. Concept 2\\
...\\

Response JSON:\\
\{\\
\hspace*{1em}"thoughts": "In this section, you explain which snippets in your text support the concepts. Be as objective as possible and ignore irrelevant information. Focus only on the snippets and avoid making guesses.",\\
\hspace*{1em}"answer": "A sentence that satisfies the concepts."\\
\}\\
\#\#\#\\

Two examples of this task being performed can be seen below. Note that the answer should be in 5 to 20 words and should be a single sentence.\\

Example 1:\\

Concepts:\\
1. The plot of the text is exciting, captivating, or engrossing. It may have unexpected twists, compelling conflicts, or keep the reader eagerly turning pages.\\
2. The characters in the movie are portrayed in a realistic and convincing manner. Their actions, dialogue, emotions, motivations, and development feel authentic and relatable, making them believable to the audience.\\
3. The narrative structure of the text is confusing or unclear, making it difficult to follow or comprehend the events happening within the story.\\
4. The text introduces some original elements or takes minor risks in the plot development, but overall, it lacks a truly unique or innovative narrative.\\

Response JSON:\\
\{\\
\hspace*{1em}"thoughts": "The snippet 'the silly and crude storyline' mentions a storyline that is described as silly and crude, indicating a lack of creativity and reliance on clichéd plot devices. The phrase 'real issues tucked between the silly and crude storyline' suggests a potentially confusing structure. It also implies real conflicts, satisfying the exciting plot concept. The term 'real issues' also supports the realistic character portrayal.",\\
\hspace*{1em}"answer": "it's about issues most adults have to face in marriage and i think that's what i liked about it -- the real issues tucked between the silly and crude storyline."\\
\}\\
\#\#\#\\

Example 2:\\
...\\
\#\#\#\\

Perform the task below, keeping in mind to limit the response to 5 to 20 words and a single sentence. Return a valid JSON response ending with \#\#\#\\

Concepts:\\
\{\}\\

Response JSON:
\end{quote}

\textbf{Image Data.}
We use the following prompt for predicate-feature mapping:
\begin{quote}
Integrate this area into the background of the image.\\
\end{quote}

\section{Empirical Evaluation (continued)}
\label{app:evelset}

\subsection{Robustness to Concept-Feature Mapping Models (Continued)}
\begin{table}
    \centering
    \caption{Explanation fidelity (accuracy) of using \toolname with different concept-feature mapping models.}

    \small
     \begin{tabular}{@{}lcccccc@{}}
        \toprule
        Model & Accuracy \\
        \midrule
        \textbf{DeepSeekV3} & 92.6 \\
        Qwen2.5 72B & 91.8 \\
        Qwen2.5 7B & 90.3\\
        \textbf{Blended Latent Diffusion} & 91.6 \\
        Latent Consistency Model & 91.4 \\
        \bottomrule
    \end{tabular}

    \label{table:exp-robust}
\end{table}

Besides the results in Section~\ref{sec:eval-robust}, we additionally conducted experiments to show the explanation fidelity of using different generative models for concept-feature mapping.
We explain a BERT model on the Large Movie Review dataset, and a ViT modle on ImageNet using \toolname local unified explanation.
Table~\ref{table:exp-robust} shows the results, which futher support that \toolname is robust to different generative model choices.


%
%

\subsection{Robustness to Prompts}
We also conducted an experiment using different prompts to generate perturbations.
We applied two other settings, one to ask GPT-4o to make our prompt template to at only 70\% of the original perturbation, and the other to ask GPT-4o to generate a prompt that is similar to the original prompt, but paraphrased.
The results are shown in Table~\ref{table:perutrbation-prompt}. We can see that the perturbation model is robust to different prompts.

\begin{table}
    \centering
    {
            \caption{Accuracy of using different prompts to generate perturbations for text data.}

    \small
     \begin{tabular}{@{}lccc@{}}
        \toprule
        Prompt & CoNLL & Onto. & Large.\\
        \midrule
        Original & 97.5 & 98.1& 97.7  \\
        Paraphrased & 97.8 & 98.0& 97.9 \\
        Shortened & 96.2 & 96.8& 96.5 \\
        \bottomrule
    \end{tabular}

    \label{table:perutrbation-prompt}
    }
\end{table}

\subsection{Robustness to Concepts}
To show the robustness of \toolname to when concepts are not ideal, we conducted two ablation studies to investigate the effects of concept absence and concept correlation on explanation fidelity using ImageNet.

\subsubsection{Concept absence}
\begin{figure}
    \color{blue}
    \centering
\resizebox{0.4\textwidth}{!}{
\includegraphics{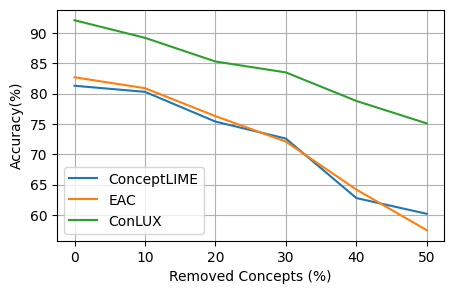}
}
\caption{Effect of concept absence on explanation fidelity.}
\label{fig:ablation-remove}
\end{figure}
When explaining YOLOv8 on ImageNet, we randomly removed a certain percentage of concepts from the concept set used for generating explanations. Figure~\ref{fig:ablation-remove} shows how explanation fidelity changes as we vary the percentage of removed concepts. The results show that \toolname exhibits greater robustness to concept removal compared to baseline concept-based explanation methods.

\subsubsection{Concept Correlation}
We conducted experiments on text data, specifically explaining BERT on the Large Movie Review dataset. In these experiments, we added a correlated concept to the concept set and measured the explanation fidelity both before and after adding the correlated concept. The results show that the accuracy slightly decreases from 92.6\% to 91.9\%, indicating that \toolname is robust to concept correlation.

\subsection{Flexibility of Supporting Various Concepts}
\label{app:more-concept}
\begin{figure}
    \color{blue}
    \centering
\resizebox{0.6\textwidth}{!}{
\includegraphics{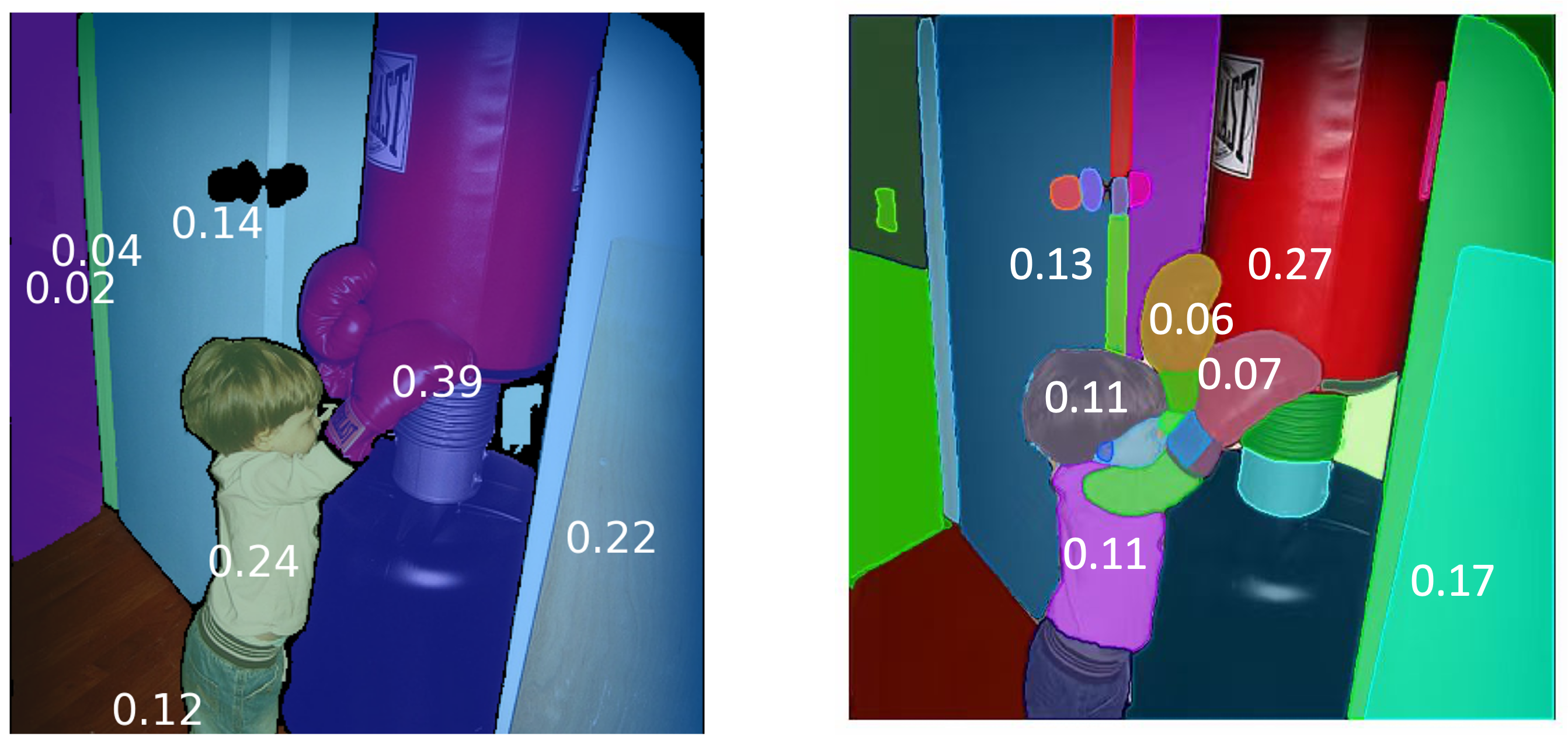}
}
\caption{\toolname-LIME explanation with hierarchical concepts.}
\label{fig:hierarchical}
\end{figure}

\toolname is a flexible framework to generate explanations with various concepts. Besides those we have shown in Section~\ref{sec:eval}, we additionally implmented \toolname for two more types of concepts to demonstrate its flexibility.
\subsubsection{Hierarchical concepts}
HIPIE~\cite{HIPIE} can extract hierarchical concepts, which can then be directly used in \toolname to produce explanations. We have implemented this, and Figure~\ref{fig:hierarchical} shows an example of hierarchical \toolname-LIME explanations.

\begin{figure}
    \color{blue}
    \centering
\resizebox{\textwidth}{!}{
\includegraphics{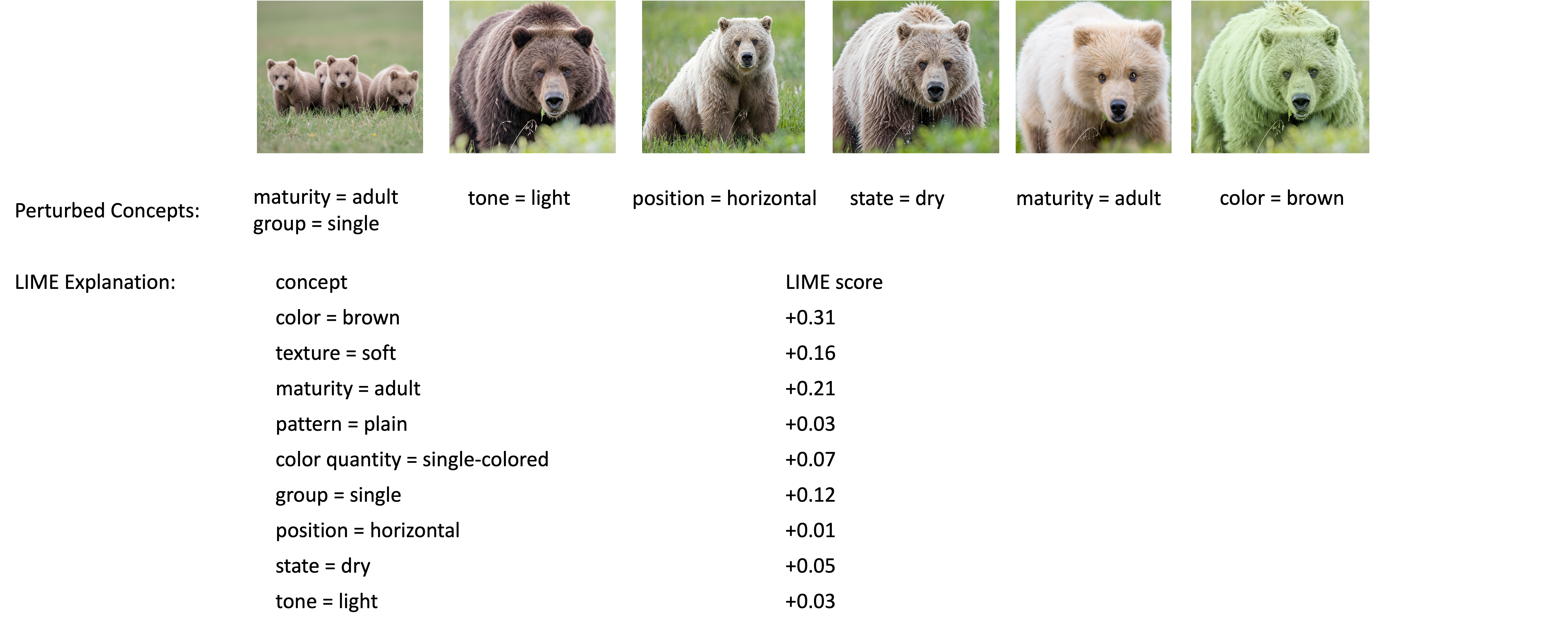}
}
\caption{\toolname-LIME perturbation samples with attributes as concepts and the corresponding explanation.}
\label{fig:attributes}
\end{figure}

\subsubsection{Attributes as Concepts}
The Open-Vocabulary Attribute Detection dataset~\cite{bravo2023open} provides attribute-level annotations (e.g., color, texture) for objects in COCO dataset. This allows \toolname to generate explanations based on these object attributes instead of replacing objects entirely. Figure~\ref{fig:attributes} shows some perturbation samples generated by \toolname-LIME using attributes as concepts and the corresponding explanation.

\subsection{Runtime Overhead (Continued)}

\begin{table}
\caption{Average runtime overhead and monetary cost of different explanation methods on image data.}
\label{table:cost-full}
\small
\centering
\begin{tabular}{lccccc}
\toprule
Method & Avg Perturbations & Avg LM Calls & Avg Time (s) & Avg Cost (\$) \\
\midrule
LIME             & 1000 & 0    & 2.35  & 0 \\
\toolname-LIME      & 1000 & 1000 & 12.31 & 0 \\
EAC              & 1000 & 0    & 7.25  & 0 \\
ConceptLIME      & 1000 & 1000 & 6.67  & 0 \\
KSHAP            & 1000 & 1000 & 2.11  & 0 \\
\toolname-KSHAP     & 1000 & 1000 & 12.45 & 0 \\
Anchors          & 193  & 0    & 3.75  & 0 \\
\toolname-Anchors   & 71   & 71   & 7.67  & 0 \\
LORE             & 851  & 0    & 6.87  & 0 \\
\toolname-LORE      & 873  & 873  & 13.45 & 0 \\
\bottomrule
\end{tabular}
\end{table}

We have discussed the computational cost of \toolname in Section~\ref{sec:eval-overhead}, 
we also provide the moneytary cost and counts the calls of the generative models in Table~\ref{table:cost-full}. 
As shown, \toolname does require additional computation time due to generative model calls. However, we note:
\begin{itemize}
    \item \textbf{Zero monetary cost:} Without compromising performance, all our experiments (Section 4 and Table 6) can use LLMs that can run on a single GPU without relying on commercial APIs, resulting in negligible monetary cost.
    \item \textbf{Acceptable overhead:} The running time of \toolname-augmented methods is practically acceptable.
\end{itemize}

\subsection{Explanation Fidelity Evaluation Details}

\subsubsection{Setup Details}
We experimented on two machines, one with an Intel i9-13900K CPU, 128 GiB RAM, and RTX 4090 GPU, and another with Intel(R) Xeon(R) Silver 4314 CPU, 256GiB RAM, and 4 RTX 4090 GPUs.

To measure the fidelity improvement brought by \toolname, we keep all hyperparameters the same for both vanilla and augmented methods.

For LIME and KSHAP, we set the number of sampled inputs to 1000 except for explaining DeepSeekV3.

For Anchors, we follow the default settings.

For LORE, we set $ngen=5$.

For the DeepSeek-V3 model, when applying it to the movie review sentiment analysis task, we simply use the following prompt:
\begin{quote}
    From now on, you should act as a sentiment analysis neural network. You should classify the sentiment of a movie review as positive or negative. If the sentence is positive, you should reply 1. Otherwise, if it's negative, you should reply 0. There may be some words that are masked in the sentence, which are represented by \textless{}UNK\textgreater{}. The input sentence may be empty, which is represented by \textless{}EMPTY\textgreater{}. You will be given the sentences to be classified, and you should reply with the sentiment of the sentence by 1 or 0.\\
There are two examples:\\
Sentence:\\
\{sentence 1 from training set\}\\
Sentiment:\\
\{label 1\}\\
Sentence:\\
\{sentence 2 from training set\} \\
Sentiment:\\
\{label 2\}\\
You must follow this format. Then I'll give you the sentence. Remember Your reply should be only 1 or 0. Do not contain any other content in your response. The input sentence may be empty.\\
Sentence:\\
\{The given sentence\}\\
Sentiment:
\end{quote}

When using Deepseek-V3 for fake news detection, we apply similar prompts to those used in other tasks. We obtain the classification probabilities by applying a softmax function to the raw probabilities of returning the two tokens: 0 and 1.

For Qwen2.5-VL, we directly prompt it to answer questions in VQAv2, and obtain the classification probabilities by applying a softmax function to the raw probabilities in Yes/No questions.

\subsubsection{Results (continued)}
In this section, we present the fidelity results that are not present in the main paper limited by space.

As Table~\ref{table:unified} shows, either \toolname-augmented KSHAP or \toolname unified explanation can outperform TBM and EAC, two state-of-the-art concept-based methods, in terms of fidelity.
We additionally provide the results of deletion experiments on TBM, EAC, and \toolname-augmented KSHAP in Table~\ref{table:concept-deletion}.

\begin{table*}[t]
    \centering
    \setlength{\tabcolsep}{1mm}
    \caption{Average accuracy (\%) (higher accuracy is better) of TBM, EAC, \toolname-augmented KSHAP (denoted as KSHAP*), and \toolname unified explanations. }
    \small
    \resizebox{0.8\textwidth}{!}{
     \begin{tabular}{@{}lcccccccccc@{}}
        \toprule
        \multirow{2}{*}{Models} & \multicolumn{5}{c}{AOPC\(\uparrow\)} & \multicolumn{5}{c}{\(\mathrm{Accuracy_{a}}(\downarrow)\)} \\
        \cmidrule(lr){2-6} \cmidrule(lr){7-11}
         & TBM & LACOAT & EAC & ConceptLIME & KSHAP*  & TBM & LACOAT & EAC & ConceptLIME & KSHAP*  \\
        \midrule
        DeepSeek-V3/Large. & 47.8 & 46.3 & -- & -- & \textbf{54.3} & 49.6 &49.8 & -- & -- & \textbf{44.1}\\
        DeepSeek-V3/Fake. & 44.2 & 45.2 & -- & -- & \textbf{51.4} & 47.4 & 48.6 & -- & -- & \textbf{45.7}\\
        BERT/Large. & 51.0 & 49.6 & -- & -- & \textbf{57.7} & 46.5 & 43.1 & -- & -- & \textbf{38.2}\\
        BERT/Fake. & 47.4 & 47.8 & -- & -- & \textbf{53.1} & 49.1 & 45.9 & -- & -- & \textbf{39.3}\\[0.2em]

        YOLOv8/ImageNet & -- & -- &52.9 & 51.1 & \textbf{62.1} & -- & -- & 6.2 & 5.9 & \textbf{5.3}\\
        YOLOv8/Caltech101 & -- & -- & 54.9 & 51.6 & \textbf{62.0} & -- & -- & 8.3 & 8.4 & \textbf{7.2}\\
        YOLOv8/CUB &--&--& 52.4 & 51.5 & \textbf{60.5} & -- & -- & 8.9 & 9.1 & \textbf{7.8} \\
        ViT/ImageNet & -- & -- & 57.1 & 54.8 & \textbf{65.4} &  -- & -- &7.1 & 6.4 & \textbf{4.5}\\
        ViT/Caltech101 & -- & -- & 62.9 & 59.4 & \textbf{69.3} & -- & -- & 11.5 & 11.9 & \textbf{10.3}\\
        ViT/CUB & -- & -- & 54.1 & 56.8 & \textbf{64.7} & -- & -- & 6.3 & 6.1 & \textbf{5.1}\\
        ResNet-50/ImageNet & -- & -- & 22.1 & 25.3 & \textbf{34.9} & -- & -- & 4.6 & 4.7 & \textbf{2.9}\\
        ResNet-50/Caltech101 & -- & -- & 41.4 & 43.6 & \textbf{53.4} & -- & -- & 7.4 & 7.8 & \textbf{5.2}\\
        ResNet-50/CUB & -- & --& 32.6 & 30.7 & \textbf{39.1} & -- & -- &5.2 & 6.0 & \textbf{4.1}\\
        \bottomrule
        \end{tabular}   
        }
    \label{table:concept-deletion}
\end{table*}

\subsection{Human Evaluation}

We introduce more detail about human evaluation in this section.

\textbf{User Screening.}
We recruited the users from students of Computer Science school in our university, and screened them before filling out the questionnaires. Therefore, the set is relatively small.
To make sure our users understood the meaning of the explanation and would predict the model output based on the explanation rather than their own feelings, we made up a test with a counter-intuitive explanation. Then we let our users predict the model outputs of ten perturbed images. We remain the users that follow the counter-intuitive explanation.

\begin{figure*}
    \centering
        \resizebox{1\textwidth}{!}{
        \includegraphics[scale=0.5]{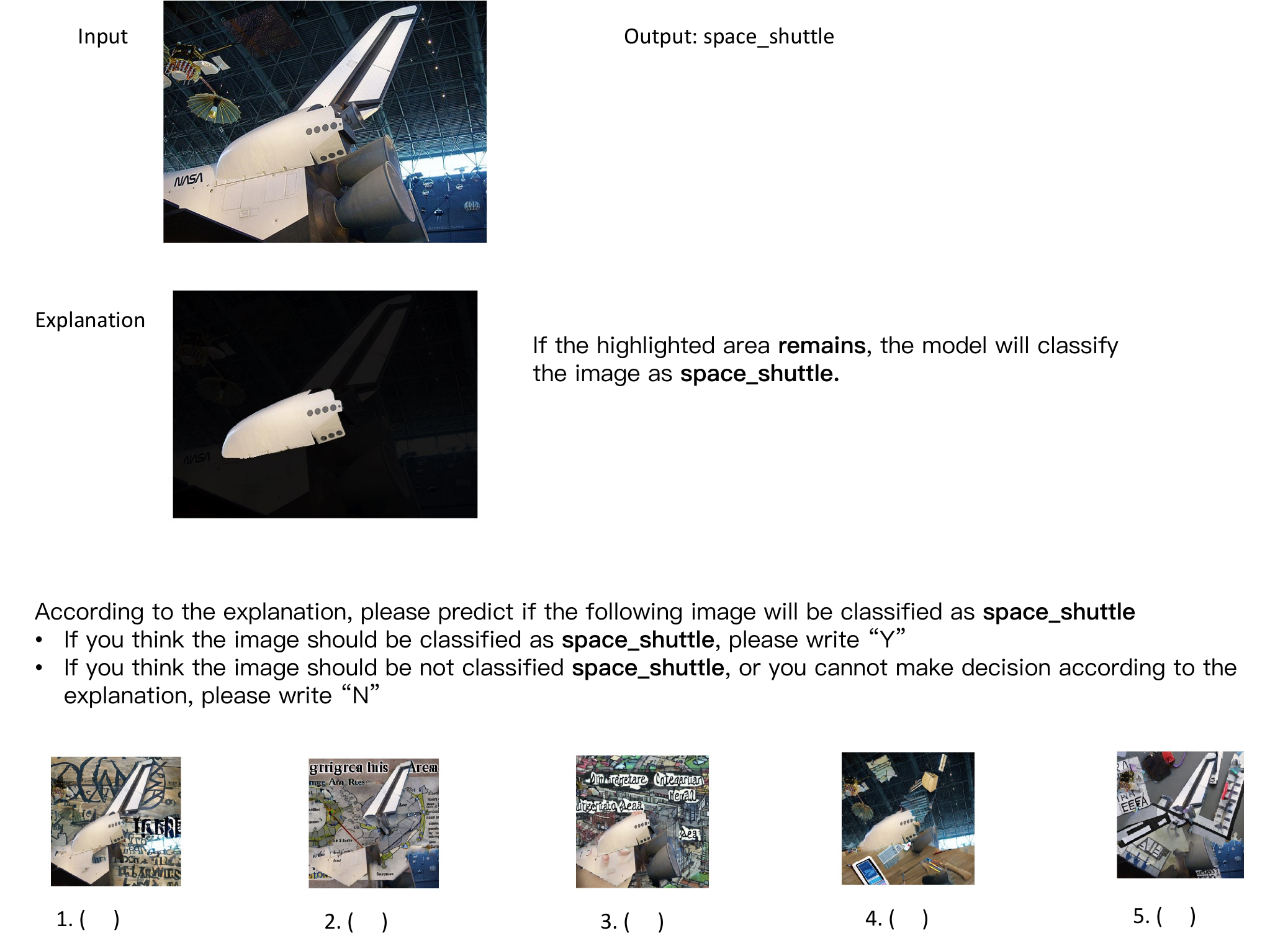} 
        }
    
\caption{An question in the task of comparing EAC and \toolname-augmented Anchors. This question provides user an \toolname-augmented Anchors explanation.}
\label{fig:usexp-anchors}
\end{figure*}

\begin{figure*}
    \centering
        \resizebox{\textwidth}{!}{
        \includegraphics[scale=0.5]{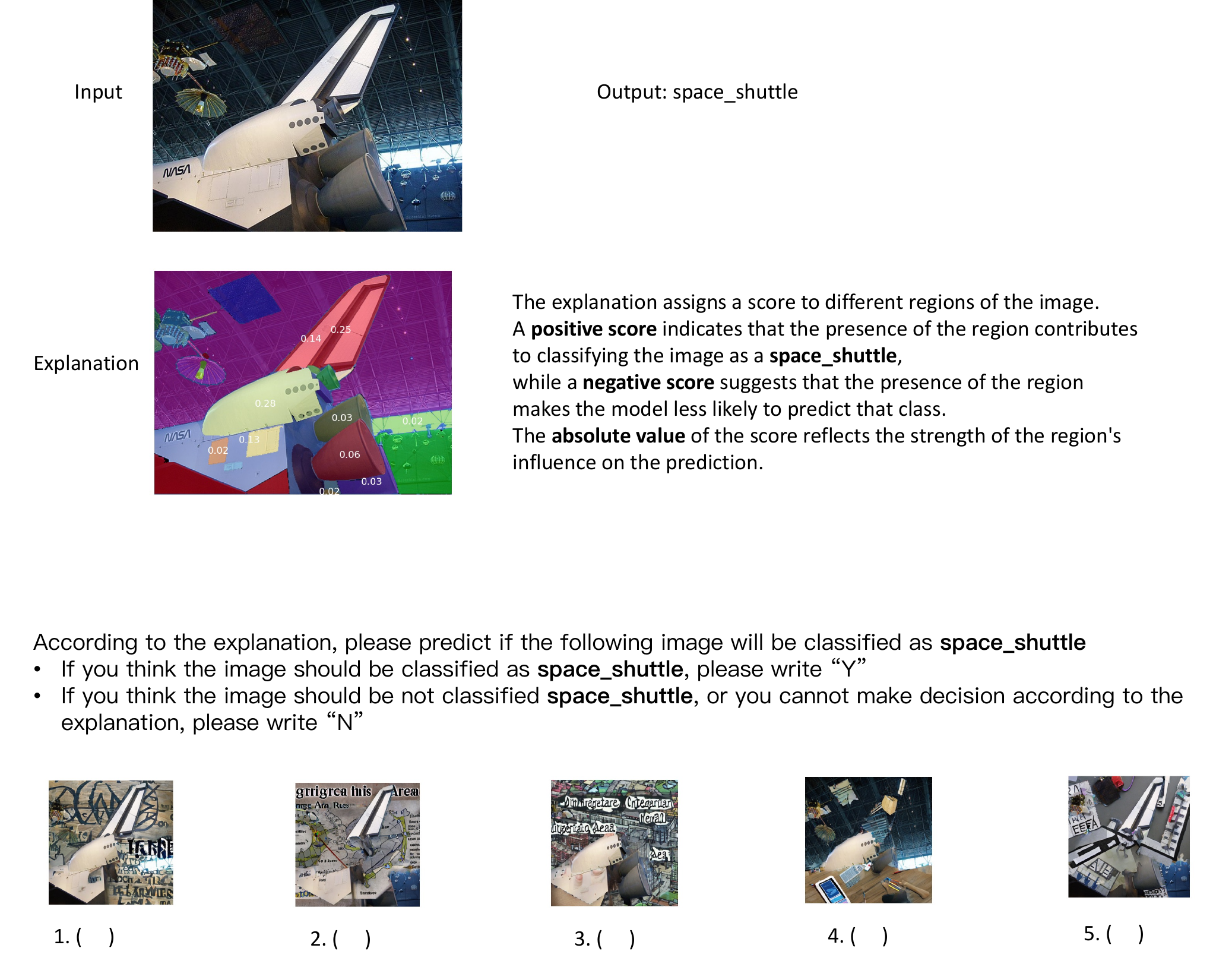} 
        }
    
\caption{A question in the task of comparing EAC and \toolname-augmented Anchors. This question provides the user with an EAC explanation.}
\label{fig:usexp-limea}
\end{figure*}

\begin{figure*}
    \centering
        \resizebox{\textwidth}{!}{
        \includegraphics[scale=0.5]{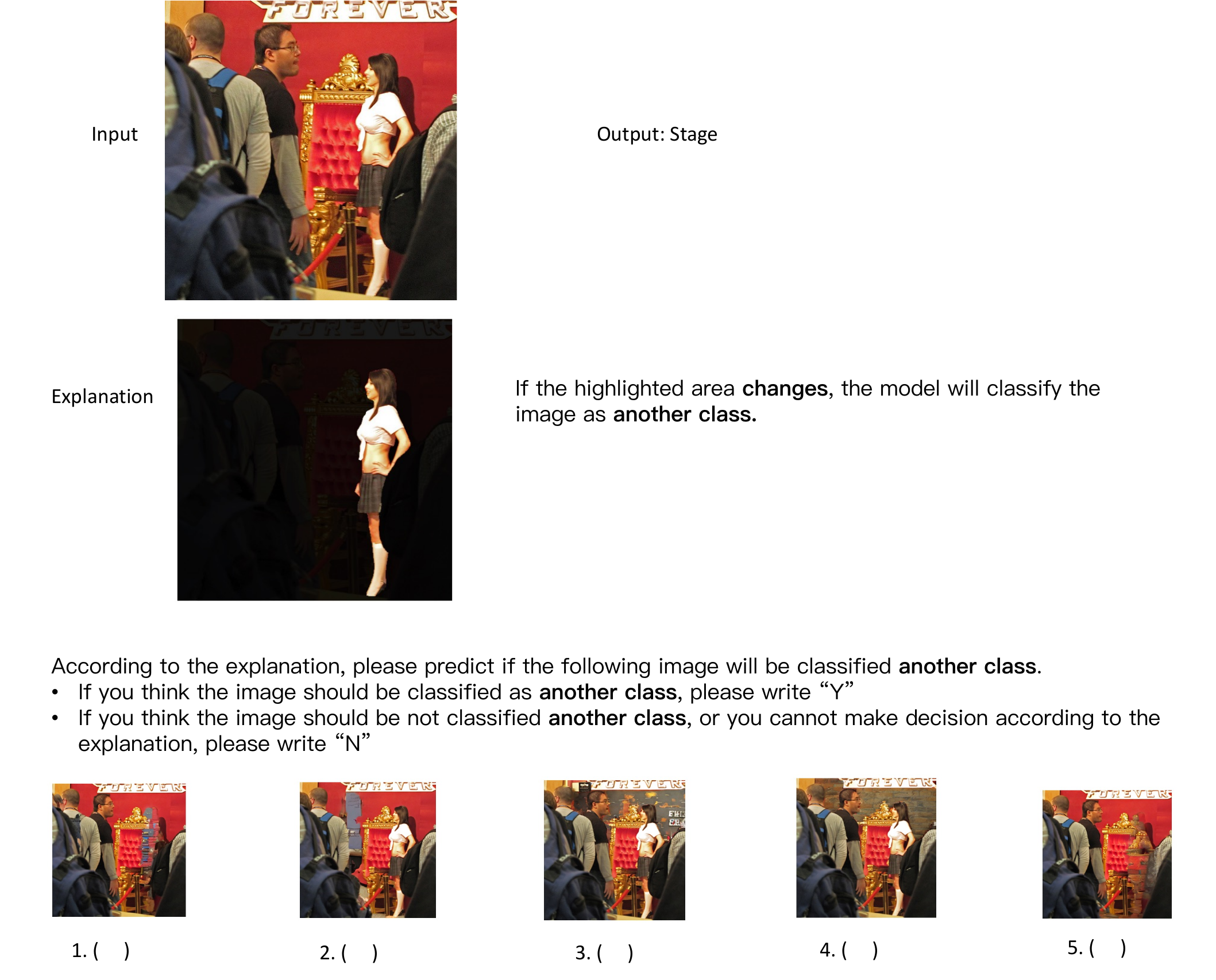} 
        }
    
\caption{A question in the task of comparing EAC and \toolname-augmented LORE. This question provides the user a \toolname-augmented LORE explanation.}
\label{fig:usexp-lore}
\end{figure*}

\begin{figure*}
    \centering
        \resizebox{\textwidth}{!}{
        \includegraphics[scale=0.5]{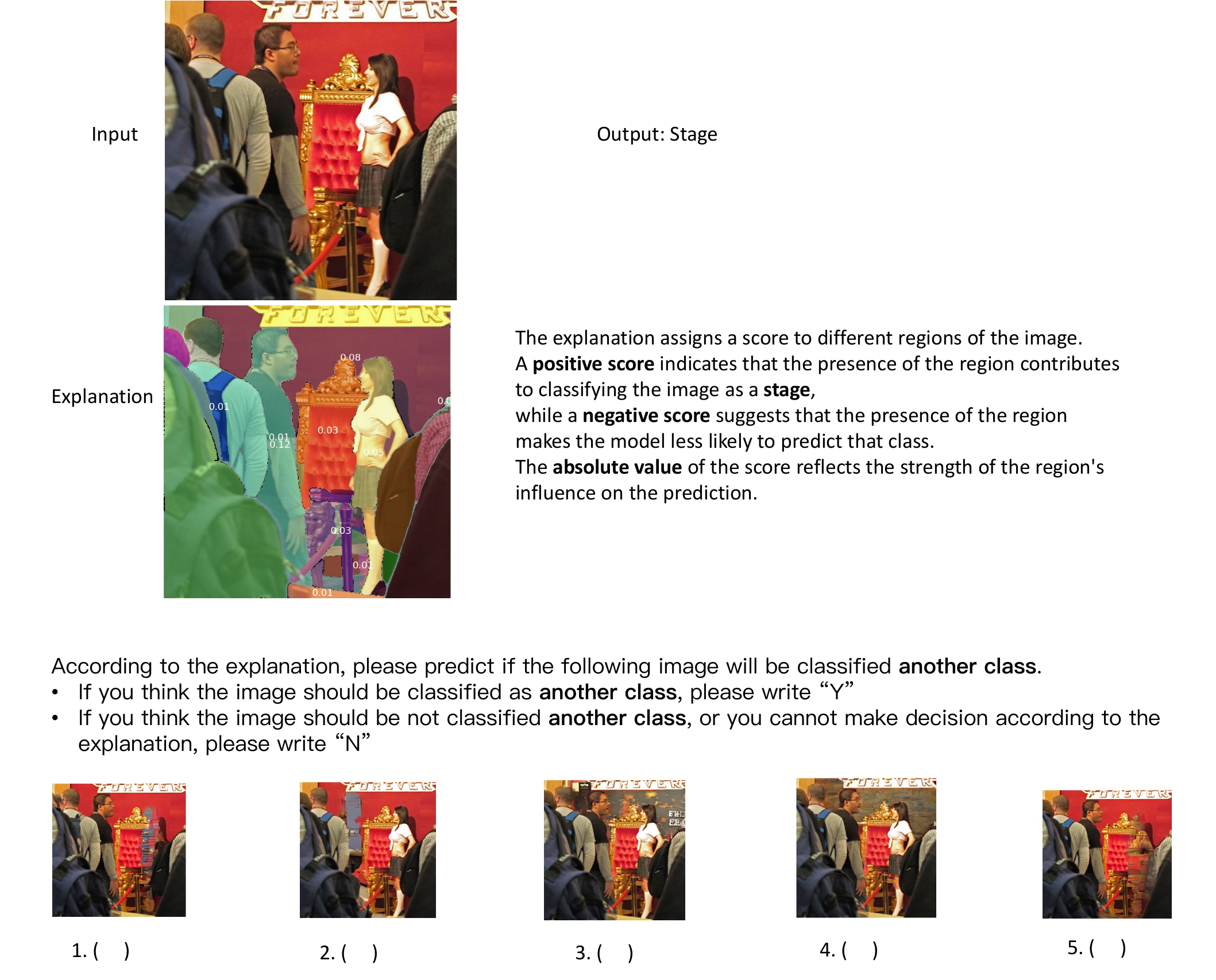} 
        }
    
\caption{An question in the task of comparing EAC and \toolname-augmented LORE. This question provides user an EAC explanation.}
\label{fig:usexp-limel}
\end{figure*}

\textbf{Questionnaires details.}
The questionnaires are similar for all users with minor variations in the order of presentation. We presented the ten questions, Q1, Q2,…, and Q10 in a random order. For each question, we presented one of the explanations generated by EAC and \toolname. For example, for the first question, user A received an explanation generated by EAC, while user B may received an explanation generated by \toolname. The users were asked to predict the model output based on the explanation.
Also, we presented the five images to be predicted in a random order.
Figure~\ref{fig:usexp-anchors}, \ref{fig:usexp-limea}, \ref{fig:usexp-lore}, and~\ref{fig:usexp-limel} show example test questions in the questionnaire.

\end{document}